\documentclass[accepted]{uai2025} % after acceptance, for a revised version;
\usepackage[american]{babel}

\usepackage{natbib}
\usepackage{bm}
\usepackage{amsthm}
\usepackage{amssymb}
\usepackage{amsmath}
\usepackage{amsfonts}
\usepackage{mathtools}
\usepackage{dsfont}

\usepackage{tikz}
\usetikzlibrary{
    calc,
    arrows.meta,
    backgrounds,
    intersections,
    shapes,
    positioning
}

\usepackage{pgfplots}
\pgfplotsset{compat=1.18}
\usepgfplotslibrary{
    groupplots,
    fillbetween
}

\usepackage{multirow}
\usepackage{enumitem}
\usepackage{balance}
\usepackage{microtype}
\usepackage{graphicx}
\usepackage{subfigure}
\usepackage{booktabs}
\usepackage{svg}
\usepackage[capitalize,noabbrev]{cleveref}
\usepackage[font=small]{caption}

\definecolor{c_link}{RGB}{1,102,94}
\definecolor{c_cite}{RGB}{191,129,45}

\usepackage{hyperref}
\hypersetup{
    colorlinks = true,
    linkcolor = {c_link},
    citecolor = {c_cite},
    urlcolor = {black},
}

\makeatletter
\def\namedlabel#1#2{\begingroup
    #2%
    \def\@currentlabel{#2}%
    \phantomsection\label{#1}\endgroup
}
\makeatother

\definecolor{c1}{RGB}{27,158,119}
\definecolor{c2}{RGB}{117,112,179}
\definecolor{c3}{RGB}{217,95,2}
\definecolor{c4}{RGB}{231,41,138}
\definecolor{c5}{RGB}{130,130,130}

\definecolor{c6}{RGB}{216,179,101}

\theoremstyle{plain}

\newtheorem{proposition}{Proposition}

\theoremstyle{definition}
\newtheorem{definition}{Definition}
\newtheorem{assumption}{Assumption}
\theoremstyle{remark}

\title{Probabilistic Graph Circuits: Deep Generative Models for Tractable\\Probabilistic Inference over Graphs}

\author[1]{\href{mailto:<papezmil@fel.cvut.cz>?Subject=PGCs}{Milan Pape\v{z}}{}}
\author[1]{Martin Rektoris}
\author[1]{V\'{a}clav \v{S}m\'{i}dl}
\author[1]{Tom\'{a}\v{s} Pevn\'{y}}

\affil[1]{%
    Artificial Intelligence Center, Czech Technical University, Prague, Czech Republic
}

\begin{document}
\maketitle

\begin{abstract}
Deep generative models (DGMs) have recently demonstrated remarkable success in capturing complex probability distributions over graphs. Although their excellent performance is attributed to powerful and scalable deep neural networks, it is, at the same time, exactly the presence of these highly non-linear transformations that makes DGMs intractable. Indeed, despite representing probability distributions, intractable DGMs deny probabilistic foundations by their inability to answer even the most basic inference queries without approximations or design choices specific to a very narrow range of queries. To address this limitation, we propose probabilistic graph circuits (PGCs), a framework of tractable DGMs that provide exact and efficient probabilistic inference over (arbitrary parts of) graphs. Nonetheless, achieving both exactness and efficiency is challenging in the permutation-invariant setting of graphs. We design PGCs that are inherently invariant and satisfy these two requirements, yet at the cost of low expressive power. Therefore, we investigate two alternative strategies to achieve the invariance: the first sacrifices the efficiency, and the second sacrifices the exactness. We demonstrate that ignoring the permutation invariance can have severe consequences in anomaly detection, and that the latter approach is competitive with, and sometimes better than, existing intractable DGMs in the context of molecular graph generation.
\end{abstract}

\section{Introduction}
\label{sec:introduction}
Graphs form a fundamental framework for modeling relations (edges) between real or abstract objects (nodes) in diverse applications, such as discovering proteins \citep{ingraham2019generative}, modeling physical systems \citep{sanchez2020learning}, detecting financial crimes \citep{li2023diga}, and searching for neural network architectures \citep{asthana2024multi}. Nonetheless, capturing the probabilistic behavior of even moderately sized graphs can be difficult. While traditional approaches struggle with this problem \citep{erdos1960evolution,holland1983stochastic,albert2002statistical}, deep generative models (DGMs) have recently proven immensely successful in this respect \citep{zhu2022survey,guo2022systematic,liu2023generative,du2024machine}.

\textbf{Challenges for graph DGMs.} There are several challenges in designing graph DGMs, including the following ones.
\begin{enumerate}[leftmargin=19pt,noitemsep,topsep=-2pt]
    \item[\namedlabel{challenges:expressiveness}{C1}.] Graphs live in large and complex combinatorial spaces. Indeed, estimated numbers of possible graphs in the molecular domain are enormous \citep{reymond2012enumeration,polishchuk2013estimation}. This poses considerable requirements on the expressivity of DGMs.
    \item[\namedlabel{challenges:random-size}{C2}.] Graphs are not random only in values of node and edge features but also in the number of these nodes and edges. Therefore, specific architectures accounting for this variable-size character of graphs are required.
    \item[\namedlabel{challenges:symmetry}{C3}.] Graphs are permutation invariant, i.e., there is a factorial number of possible configurations of a single graph. The key property of graph DGMs should be to recognize all the configurations as the same graph.
    \item[\namedlabel{challenges:validity}{C4}.] Graphs respect domain-specific semantic validity. For example, not all molecular graphs are chemically valid but must adhere to chemical valency constraints.
\end{enumerate}

\textbf{Current solutions.} Depending on the mechanism of generating their outputs, graph DGMs can broadly be divided into \emph{autoregressive} and \emph{one-shot} models. To ensure sufficient expressiveness (\ref{challenges:expressiveness}), both these categories rely on graph neural networks (GNNs) \citep{wu2020comprehensive,zhang2020deep}.
% which prove successful despite limitations following from the $k$-Weisfeiler-Leman isomorphism test \citep{morris2023weisfeiler}.
To accommodate for the varying size (\ref{challenges:random-size}), autoregressive models treat graphs as sequences whose elements are recursively processed by recurrent units \citep{liao2019efficient}, whereas one-shot models use virtual-node padding \citep{madhawa2019graphnvp} or convolutional GNNs \citep{de2018molgan}. The permutation invariance (\ref{challenges:symmetry}) can be imposed in various ways \citep{murphy2019relational}. Two techniques stand up frequently for both the categories of DGMs. The first relies on sorting the graph into its canonical configuration \citep{chen2021order}, and the second utilizes the permutation equivariance of graph neural networks \citep{niu2020permutation}. To capture the semantic validity (\ref{challenges:validity}), both the categories can benefit from domain knowledge, e.g., by using rejection sampling \citep{shi2020graphaf}, post-hoc correction \citep{zang2020moflow}, constraints \citep{liu2018constrained}, or regularization \citep{ma2018constrained}. However, it is preferable to satisfy the validity in a completely domain-agnostic way \citep{zang2020moflow}, relying on the model's expressiveness to extract the essence of the problem.

\textbf{The limitation of graph DGMs.} Graph DGMs excel at sampling new graphs, as it is their predetermined probabilistic inference task. However, despite constituting models of a probability distribution, other inference tasks---such as marginalization, conditioning, or expectation---remain elusive. In other words, graph DGMs are \emph{intractable} probabilistic models. There are a few exceptions to that, allowing for one or two more inference tasks beyond mere sampling. Graph normalizing flows \citep{madhawa2019graphnvp} and graph autoregressive models \citep{you2018graphrnn} allow for exact likelihood evaluation (conditionally on the canonical configuration of a graph). The autoregressive models additionally provide a narrow range of marginal and conditional distributions, only those respecting the fixed sequential structure of the chain rule of probability. Alternatively, it is a common practice to equip DGMs with extra components explicitly tailored to desired inference tasks, such as conditional sampling \citep{vignac2023digress}.

\textbf{Tractable probabilistic models.} Probabilistic circuits~(PCs) constitute a framework \citep{choi2020probabilistic} which unifies many tractable probabilistic models (TPMs) as special cases. PCs are expressive (\ref{challenges:expressiveness}) deep networks \citep{martens2015expressive,de2021compilation,yin2024expressive} that can encode probability distributions with hundreds of millions of parameters \citep{liu2023scaling,liu2023understanding,liu2024scaling}. The underlying characteristic of PCs is that they provide exact and efficient answers to a wide range of inference queries---such as marginalization, conditioning, and expectation---without approximations or query-specific architecture modifications. However, their adoption to graphs has received limited attention. Existing PCs for graphs address \ref{challenges:random-size} both in (pseudo-)autoregressive \citep{errica2024tractable} or one-shot \citep{nath2015learning,papez2024sum} manner, adopting the chain rule of probability with strong independence assumption, whereas other models treat graphs as fixed-size objects \citep{zheng2018learning,ahmed2022semantic,loconte2023turn}. To the best of our knowledge, there is currently no PC for graphs satisfying \ref{challenges:symmetry}, with the exceptions \citep{nath2015learning,papez2024sum} that are only partially permutation invariant \citep{diaconis1984partial}. These approaches are, however, designed only for specific types of graphs. Moreover, PCs are well-suited to inject domain-specific validity constraints (\ref{challenges:validity}) through knowledge compilation \citep{ahmed2022semantic,loconte2023turn}. We offer more remarks on the related work in \cref{sec:related-work}.

\textbf{Contributions.} The contributions of this paper are summarized as follows.
\begin{itemize}[leftmargin=9pt,noitemsep,topsep=-2pt]
    \item To address the limitations of the aforementioned DGMs, we propose probabilistic graph circuits (PGCs), a framework for designing one-shot graph TPMs (that can operate also in the autoregressive regime due to their tractability).
    \item We extend the standard notion of permutation invariance---which can involve intractable distributions---to a more strict form of \emph{tractable} permutation invariance (\ref{challenges:symmetry}) and specify the conditions for the tractability of PGCs.
    \item To address the variable-size character of graphs (\ref{challenges:random-size}), we utilize the key benefit of PCs---i.e., their tractability---and propose the \emph{marginalization padding}, which is an approach that marginalizes out non-existing nodes (and associated edges) of input graphs.
    \item We define conditions for making PGCs inherently and tractably permutation invariant (\ref{challenges:symmetry}) and discuss that achieving such property can lead to reduced expressive power. Therefore, we investigate two principles that make PGCs permutation invariant even with arbitrary permutation-sensitive building blocks: the first ensures invariance through marginalization, and the second through conditioning on a canonical graph ordering.
    \item We illustrate the importance of the permutation invariance in a synthetic anomaly detection example. Then, we show that the order-conditioned PGCs (which we implement using several state-of-the-art variants of PCs and different canonical graph orderings) are competitive---and in most metrics even superior---to intractable graph DGMs in the context of the unconditional generation of molecular graphs. To demonstrate that the PGCs can perform tractable inference tasks efficiently, we generate molecular graphs conditionally on known subgraphs.
\end{itemize}

\begin{figure*}[ht]
\begin{center}
\centerline{\input{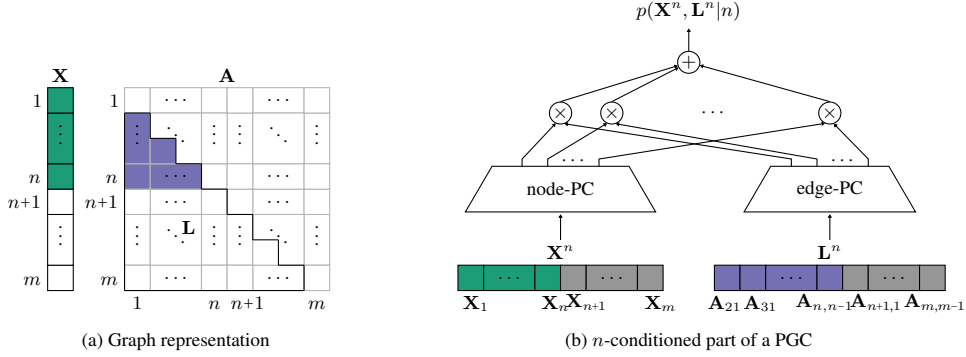}}
\vspace{-5pt}
\caption{\emph{An example of a PGC for undirected acyclic graphs.} (a) We consider a graph $\mathbf{G}$ represented by a feature matrix, $\mathbf{X}$, and an adjacency tensor, $\mathbf{A}$, such that each instance of $\mathbf{G}$ (highlighted in \textcolor{c1}{green} and \textcolor{c2}{blue}) has a random number of nodes, $n\in(0,1,\ldots,m)$, where $m$ is a fixed maximum number of nodes. The empty places (white) are not included in the training data. (b) The main building block of PGCs is the $n$-conditioned joint distribution over $\mathbf{X}^n$ and $\mathbf{L}^n$, the latter of which is a flattened lower triangular part of $\mathbf{A}^n$. $\mathbf{X}^n$ and $\mathbf{L}^n$ are used as input into the node-PC and edge-PC, respectively. The empty places are marginalized out (\textcolor{c5}{grey}). The outputs of these two PCs are passed through the product layer with $n_c$ units and the sum layer with a single unit.
}
\vspace{-30pt}
\label{fig:pgc}
\end{center}
\end{figure*}

\vspace{-4pt}
\section{Preliminaries}\label{sec:preliminaries}
\vspace{-2pt}

\textbf{Notation.} We denote random variables by upper-case~letters, $X$, and their realizations by lower-case letters, $x\in\mathsf{dom}(X)$, where $\mathsf{dom}(X)$ is the domain of~$X$. Sets of random variables are expressed by bold upper-case letters, $\mathbf{X}\coloneqq\lbrace X_1,\ldots,X_n\rbrace$, and their realizations by bold lower-case letters, $\mathbf{x}\coloneqq\lbrace x_1,\ldots,x_n\rbrace$. The domain of $\mathbf{X}$ is $\mathsf{dom}(\mathbf{X})\coloneqq\mathsf{dom}(X_1)\times\cdots\times\mathsf{dom}(X_n)$. To denote the set of positive integers, we use $[n]\coloneqq\lbrace 1,\ldots,n\rbrace$ with $n>0$. If $\mathsf{dom}(X_i)$ is identical for all $i\in [n]$, we write $\mathsf{dom}(\mathbf{X})\coloneqq\mathsf{dom}(X)^n$. We often consider that the elements of $\mathbf{X}$ are organized into a matrix or a tensor such that $\mathsf{dom}(\mathbf{X})\coloneqq\mathsf{dom}(X)^{n_1\times n_2}$ or $\mathsf{dom}(\mathbf{X})\coloneqq\mathsf{dom}(X)^{n_1\times n_2\times n_3}$, respectively. To index $(i,j,k)$-th entry of a tensor, we use subscripts, i.e., $\mathbf{X}_{ijk}$. To select slices of a tensor, we use the colon ``:'', e.g., $\mathbf{X}_{::k}$ is the $k$-th matrix along the third dimension of $\mathbf{X}$. We often use $\mathbf{X}_{i}\coloneqq\mathbf{X}_{i::}$ or $\mathbf{X}_{ij}\coloneqq\mathbf{X}_{ij:}$ to abbreviate the notation.

Importantly, we consider that all sets of variables are random not only in their values but also in their size, i.e., in $\mathbf{X}\coloneqq\lbrace X_1,\ldots,X_N\rbrace$, not only each $X_i$ is random but $N$ is also random, which we sometimes stress by writing $\mathbf{X}\coloneqq\mathbf{X}^N$. When a set is random only in values but fixed in size, we use $\mathbf{X}^n$. For a realization of values and size, we write $\mathbf{x}\coloneqq\mathbf{x}^n$. The stochastic behavior of a fixed-size random set, $\mathbf{X}^n$, is characterized by a probability distribution, $p(\mathbf{X}^n|n)$, whose structure and the number of parameters depends on $n$. We make this dependence deliberately explicit in this paper. On the contrary, $p(\mathbf{X})\coloneqq p(\mathbf{X}^N,N)$ is a probability distribution over random values and random~size.

\textbf{Probabilistic circuits.} PCs are deep computational networks \citep{van2019tractable,vergari2019visualizing} composed of three types of computational units: input, sum, and product units \footnote{We refer the reader to \cref{sec:circuits} for an introduction to PCs.}. The key property of PCs is that---under structural constraints \citep{darwiche2002knowledge}---they are \emph{tractable}, as described in \cref{def:tractability-pc}.
\begin{definition}{(Tractability of PCs \citep{choi2020probabilistic}).}\label{def:tractability-pc}
    Let $p(\mathbf{X}^n|n)$ be a PC encoding a probability distribution over a fixed-size set, $\mathbf{X}^n$, which is smooth (\cref{ass:smoothness}), decomposable (\cref{ass:decomposability}), and has tractable input units (\cref{ass:tractable-input-units}). Furthermore, consider that $\mathbf{X}^n$ can be organized into two fixed-size subsets, $\mathbf{X}^n\coloneqq\lbrace\mathbf{X}_a^{n-k},\mathbf{X}_b^k\rbrace$. Then, the integral $\int p(\mathbf{x}_a^{n-k},\mathbf{X}_b^k|n)d\mathbf{x}_a^{n-k}$ can be computed (\textbf{A}) \emph{exactly} (without approximations) and (\textbf{B}) \emph{efficiently} with $\mathcal{O}(\text{poly}(|p|))$ complexity, where $|p|$ is the number of connections between the computational units of $p$.
\end{definition}
\cref{def:tractability-pc} is the key distinguishing feature of PCs. Its main consequence is that many inference tasks \citep{vergari2021compositional,wang2024compositional} can be performed in a single forward pass through the network.

\textbf{Graphs.} We define an $N$-node graph as a tuple, $\mathbf{G}\coloneqq\lbrace\mathbf{X},\mathbf{A}\rbrace$, containing a node feature matrix, $\mathbf{X}$, and an edge adjacency tensor, $\mathbf{A}$. The domains $\mathsf{dom}(\mathbf{X})$ and $\mathsf{dom}(\mathbf{A})$ are defined by $\mathsf{dom}(X)^{N\times n_X}$ and $\mathsf{dom}(A)^{N\times N\times n_A}$, respectively (where $N$ is random but $n_X$ and $n_A$ are always fixed). To express $\mathbf{G}$ with $n_X$ types of nodes and $n_A$ types of edges, we consider that $\mathsf{dom}(\mathbf{X})$ and $\mathsf{dom}(\mathbf{A})$ are one-hot encoded along the last (fixed-size) dimension. Therefore, it holds that $\mathsf{dom}(X)\coloneqq\lbrace 0,1\rbrace$, $\sum^{n_X}_{j=1}\mathbf{X}_{ij}=1$, and $\mathsf{dom}(A)\coloneqq\lbrace 0,1\rbrace$, $\sum^{n_A}_{k=1}\mathbf{A}_{ijk}=1$. We assume that $\mathsf{dom}(\mathbf{A})$ includes an extra category to express that there can be no connection between two nodes.
% Note that $\mathsf{dom}(X)$ and $\mathsf{dom}(A)$ can easily be extended to accommodate continuous features, which we do not consider in this work. 

\textbf{Graphs are $\mathbb{S}_n$-invariant.} There can be up to $n!$ distinct but equivalent permutations (orderings) of an $n$-node graph, $\mathbf{G}^n$. A properly designed probabilistic model, $p(\mathbf{G}^n|n)$, has to recognize all these configurations by assigning them with the same probability. This property is known as permutation invariance or $\mathbb{S}_n$-invariance. To define $\mathbb{S}_n$-invariance of a probability distribution over a graph, $p(\mathbf{G}^n|n)$, we use a finite symmetric group of a set of $n$ elements, $\mathbb{S}_n$. This is a set of all $n!$ permutations of $[n]$. We consider that each permutation, $\bm{\pi}\in\mathbb{S}_n$, acts upon the first dimension of the feature matrix, $\bm{\pi}\mathbf{X}^n\coloneqq\lbrace\mathbf{X}_{\pi(1)},\ldots,\mathbf{X}_{\pi(n)}\rbrace$, and upon the first two dimensions of the adjacency tensor, $\bm{\pi}\mathbf{A}^n\coloneqq\lbrace\mathbf{A}_{\pi(1)\pi(1)},$ $\mathbf{A}_{\pi(1)\pi(2)},\ldots,\mathbf{A}_{\pi(n)\pi(n)}\rbrace$ \citep{orbanz2014bayesian}. This allows us to permute an $n$-node graph, $\mathbf{G}^n\coloneqq(\mathbf{X}^n,\mathbf{A}^n)$, as follows: $\bm{\pi}\mathbf{G}^n=(\bm{\pi}\mathbf{X}^n,\bm{\pi}\mathbf{A}^n)$. We formally introduce the $\mathbb{S}_n$-invariance of $p(\mathbf{G}^n|n)$ in \cref{def:sn-graph-invariance}.

\begin{definition}{($\mathbb{S}_n$-invariance).}\label{def:sn-graph-invariance}
    The probability distribution $p(\mathbf{G}^n|n)$ is $\mathbb{S}_n$-invariant iff $p(\bm{\pi}\mathbf{G}^n|n)=p(\mathbf{G}^n|n)$ for all $\bm{\pi}\in\mathbb{S}_n$, and, therefore, $\mathbf{G}$ is $\mathbb{S}_n$-invariant if $p(\mathbf{G}^n|n)$~is.
\end{definition}%

\textbf{Problem definition.} We aim to design a \emph{tractable}, $\mathbb{S}_n$-\emph{invariant} probability distribution over graphs, $p(\mathbf{G})$, and learn its parameters based on a collection of observed graphs, $\lbrace\mathbf{G}_1,\ldots,\mathbf{G}_I\rbrace$, where each $\mathbf{G}_i$ can have a different number of nodes and edges.

\vspace{-4pt}
\section{Probabilistic Graph Circuits}\label{sec:pgcs}
\vspace{-2pt}

Conventional PCs encode a tractable probability distribution over a fixed-size set, $p(\mathbf{X}^n|n)$; however, they are not $\mathbb{S}_n$-invariant \citep{papez2024sum}. In contrast, we propose probabilistic graph circuits that encode a probability distribution over a random-size graph, $p(\mathbf{G})$, considering different ways to ensure their $\mathbb{S}_n$-invariance and its impact on the tractability. We aim that $p(\mathbf{G})$ inherits the ability to answer the broad range of probabilistic inference queries from conventional PCs. To introduce (probabilistic) graph circuits, we start with defining the scope of a graph, which will be important in describing their inner mechanisms.

\begin{definition}[Graph Scope]\label{def:graph-scope}
    The scope of $\mathbf{G}$ is an arbitrary subset, $\mathbf{G}_u\subseteq\mathbf{G}$, such that $\mathbf{X}_u\subseteq\mathbf{X}$ and $\mathbf{A}_u\subseteq\mathbf{A}$, i.e., it can be a subgraph, $\mathbf{G}_u=(\mathbf{X}_u,\mathbf{A}_u)$, a set of nodes, $\mathbf{G}_u=\mathbf{X}_u$, or a set of edges, $\mathbf{G}_u=\mathbf{A}_u$.
\end{definition}

Note that if we split $\mathbf{G}$ into two scopes $\mathbf{G}\coloneqq\lbrace \mathbf{G}_a, \mathbf{G}_b\rbrace$, then both $\mathbf{G}_a$ and $\mathbf{G}_b$ have a random number of nodes such that $N=N_a+N_b$.

Graph circuits are semantically similar to conventional circuits in that they comprise sum, product, and input units. However, the key difference is that these units are defined over variable-size subgraphs rather than fixed-size subsets, as introduced in Definitions \ref{def:graph-circuit} and \ref{def:probabilistic-graph-circuit}.

\begin{definition}[Graph Circuit]\label{def:graph-circuit}
    A \emph{graph circuit} (GC) $c$ over a random graph $\mathbf{G}\coloneqq(\mathbf{X},\mathbf{A})$ is a parameterized computational network encoding a function $c(\mathbf{G})$. It contains three types of computational \emph{units}: \emph{input}, \emph{product}, and \emph{sum} units. All sum and product units receive the outputs of other units as inputs. We denote the set of inputs of a unit $u$ as $\mathsf{in}(u)$. Each unit $u$ encodes a function $c_u$ over a subgraph, $\mathbf{G}_u\subseteq\mathbf{G}$, the \emph{graph scope} (\cref{def:graph-scope}). The input unit $c_u(\mathbf{G}_u)\coloneqq f_u(\mathbf{G}_u)$ computes a user-defined, parameterized function, $f_u$; the sum unit computes the weighted sum of its inputs, $c_u(\mathbf{G}_u)\coloneqq\sum_{i\in\mathsf{in}(u)}w_ic_i(\mathbf{G}_i)$, where $w_i\in\mathbb{R}$ are the weight parameters; and the product unit computes the product of its inputs, $c_u(\mathbf{G}_u)\coloneqq\prod_{i\in\mathsf{in}(u)}c_i(\mathbf{G}_i)$. The scope of any sum or product unit is the union of its input scopes, $\mathbf{G}_u=\bigcup_{i\in\mathsf{in}(u)}\mathbf{G}_i$\footnote{For example, making the random number of nodes of $\mathbf{G}$ explicit, the union of scopes for a product unit with two children is $\mathbf{G}^N=\mathbf{G}_a^{N_a}\cup\mathbf{G}_b^{N_b}$, where $N=N_a+N_b$.}. A circuit can have one or multiple root units. The scope of a root unit is $\mathbf{G}$.
\end{definition}%

\begin{definition}[Probabilistic Graph Circuit]\label{def:probabilistic-graph-circuit}
    A \emph{probabilistic graph circuit} (PGC) over a random graph $\mathbf{G}$ is a GC (\cref{def:graph-circuit}) $c$, such that $\forall \mathbf{g}\in\mathsf{dom}(\mathbf{G}): c(\mathbf{g})\geq 0$, i.e., it is a \emph{non-negative} function for all realizations of $\mathbf{G}$.
\end{definition}

A PGC (\cref{def:probabilistic-graph-circuit}) encodes a possibly unnormalized, joint probability distribution over the size and values of $\mathbf{G}$. \cref{def:graph-scope} implies that the scope of each computational unit of a (P)GC has a random number of nodes (and edges). This random character of computational units opens up various constructions of PGCs. In this paper, we start to investigate PGCs that take the following form:
\begin{equation}\label{eq:pgc}
    p(\mathbf{G})=p(\mathbf{G}^N,N)=p(\mathbf{G}^n|n)p(N),
\end{equation}
where $p(\mathbf{G}^n|n)$ is an $n$-conditioned PGC over an $n$-node graph, $\mathbf{G}^n$, which is fixed in its size but describes stochastic behavior of values that the nodes and edges of $\mathbf{G}$ can take; and $p(N)$ is a cardinality distribution, which characterizes the random size of~$\mathbf{G}$.

\vspace{-4pt}
\subsection{Tractability of PGCs}
\vspace{-2pt}
One of the key requirements for GCs is that they should satisfy $\mathbb{S}_n$-invariance (\ref{challenges:symmetry}). However, the standard definition of $\mathbb{S}_n$-invariance (\cref{def:sn-graph-invariance}) is too general. Namely, it does not specify how $p$ achieves this property and, especially, whether $p$ is tractable. Tractability of $\mathbb{S}_n$-invariant functions in the deep learning literature takes only the perspective of the computational efficiency \citep{murphy2019relational}. Therefore, we extend \cref{def:sn-graph-invariance} with both parts (A and B) of \cref{def:tractability-pc} that are canonical in the circuit literature. The PGC \eqref{eq:pgc} is $\mathbb{S}_n$-invariant only if its $n$-conditioned part is $\mathbb{S}_n$-invariant. We thus describe the tractable $\mathbb{S}_n$-invariance in terms of $p(\cdot|n)$, as provided in \cref{def:tractability-sn-invariance}.

\begin{definition}{(Tractable $\mathbb{S}_k$-invariance.)}\label{def:tractability-sn-invariance}
    Let $p(\mathbf{G}^n|n)$ be an $n$-conditioned, smooth (\cref{ass:smoothness}) and decomposable (\cref{ass:decomposability}) PGC whose input units admit tractable integration (\cref{ass:tractable-input-units}). Furthermore, consider that $\mathbf{G}^{n}$ is organized into two \emph{fixed-size} subgraphs, $\mathbf{G}^n=\lbrace\mathbf{G}^{n-k}_a,\mathbf{G}^k_b\rbrace$. Then, $p(\cdot|n)$ is \emph{tractably} $\mathbb{S}_k$-invariant~if
    \begin{equation}\nonumber
        \int p(\mathbf{g}^{n-k}_a,\bm{\pi}^{}_b\mathbf{G}^k_b|n)d\mathbf{g}^{n-k}_a=\int p(\mathbf{g}^{n-k}_a,\mathbf{G}^k_b|n)d\mathbf{g}^{n-k}_a
    \end{equation}
    can be computed \emph{exactly} in $\mathcal{O}(\text{poly}(|p|))$ time for all $\bm{\pi}_b\in\mathbb{S}_{k}$, where $d\mathbf{g}^{n-k}_a$ is a suitable reference measure corresponding to the $(n\text{-}k)$-node realization~$\mathbf{g}^{n-k}_a$.
\end{definition}

Tractable $\mathbb{S}_k$-invariance results in tractable $\mathbb{S}_n$-invariance for $k=n$. Although there is then no integral to compute in \cref{def:tractability-sn-invariance}, this case implies that the full evidence query $p(\mathbf{g}^n|n)$ is obtained \emph{exactly} in $\mathcal{O}(\text{poly}(|c|))$ time. In other words, we obtain the $p(\mathbf{g}^n|n)$ value in a single forward pass through the circuit. Under this $k=n$ setting, we are backward compatible to \cref{def:sn-graph-invariance}, yet now we specify what tractability means in this context.

The tractable $\mathbb{S}_k$-invariance (\cref{def:tractability-sn-invariance}) is the key element in the tractability of PGCs. However, the tractability of \eqref{eq:pgc} does not depend only on the $n$-conditioned part, $p(\cdot|n)$, but also on the cardinality distribution, $p(N)$, as specified in \cref{prop:tractability-pgc}.

\begin{proposition}{(Tractability of PGCs.)}\label{prop:tractability-pgc}
    Let $p$ be a PGC \eqref{eq:pgc} such that $p(\mathbf{G}^n|n)$ is tractably $\mathbb{S}_n$-invariant (\cref{def:tractability-sn-invariance}), and $p(N)$ has a finite support. Furthermore, consider that $\mathbf{G}$ is organized into two \emph{random-size} subgraphs, $\mathbf{G}=\lbrace\mathbf{G}_a,\mathbf{G}_b\rbrace$. Then, $p(\mathbf{G})$ is tractable if
    \begin{equation}\nonumber
        \int p(\mathbf{g}_a,\mathbf{G}_b)d\mathbf{g}_a
        =
        \sum^\infty_{n=k}\int p(\mathbf{g}^{n-k}_a,\mathbf{G}^k_b|n)p(n)d\mathbf{g}^{n-k}_a
    \end{equation}
    can be computed \emph{exactly} in $\mathcal{O}(\text{poly}(|p|))$ time.
\end{proposition}

\cref{prop:tractability-pgc} invokes an impression that the infinite sum makes the integral intractable. However, $p(n)$ is assumed to have finite support, i.e., it is always occupied with non-zero values only in a specific interval, $0<n\leq m$, while being equal to zero outside. For this reason, the infinite sum becomes a finite one, and the integral is tractable.

\textbf{Marginalization padding.} The conventional PCs model probability distributions over fixed-size vectors. However, each instance of $\mathbf{G}$ can have a different number of nodes and edges (\ref{challenges:random-size}). The PGC \eqref{eq:pgc} partially reflects the variable-size character of $\mathbf{G}$ by $p(N)$, which describes the random size of the $N$-node graph, $\mathbf{G}^N$. Still, the $n$-node (fixed-size) argument $\mathbf{G}^n$ in $p(\cdot|n)$ raises the question: how to design $p(\cdot|n)$---which should be characterized by a single, fixed-size set of parameters---such that it can take graphs with different values of $n$ as input (e.g., such that we can easily enumerate the sum terms in \cref{prop:tractability-pgc})? To solve this problem, we utilize the key feature of PGC---their tractability---to propose \emph{marginalization padding}. This mechanism assumes that there can be at most $m$ nodes in a random-size graph, $\mathbf{G}$, and whenever there is an instance with $n<m$ nodes, it marginalizes out the remaining $m-n$ `empty' nodes and associated edges. We illustrate this principle on a specific instance of PGCs in \cref{fig:pgc}.

\vspace{-4pt}
\subsection{Inherent $\mathbb{S}_n$-invariance}\label{sec:sn-invariance-inherent}
\vspace{-2pt}
There are different ways to design a PGC that satisfies \cref{def:tractability-sn-invariance}. The primary question we ask is how to design a PGC that is \emph{inherently} $\mathbb{S}_n$-invariant? The inherent $\mathbb{S}_n$-invariance means that a PGC satisfying Assumptions \ref{ass:smoothness}-\ref{ass:tractable-input-units} is tractably $\mathbb{S}_n$-invariant without resorting to any external $\mathbb{S}_n$-invariance mechanism. The following definition states the conditions for such a PGC.
\begin{definition}{(Inherently $\mathbb{S}_n$-invariant PGCs.)}\label{def:conditions-sn-invariance}
    A tractable PGC (\cref{prop:tractability-pgc}) is inherently $\mathbb{S}_n$-invariant if its computational units satisfy the following conditions:
    \begin{itemize}[leftmargin=19pt,noitemsep,topsep=-2pt]
        \item[a)] $p_u(\bm{\pi}_u\mathbf{G}_u)\coloneqq f_u(\bm{\pi}_u\mathbf{G}_u)$,
        \item[b)] $p_u(\bm{\pi}_u\mathbf{G}_u)\hspace{2pt}=\sum_{i\in\mathsf{in}(u)}w_ip_i(\bm{\pi}_u\mathbf{G}_i)$,
        \item[c)] $p_u(\bm{\pi}_u\mathbf{G}_u)\hspace{2pt}=\prod_{i\in\mathsf{in}(u)}p_i(\bm{\pi}_i\mathbf{G}_i)$,
    \end{itemize}
    for all $\bm{\pi}_u\subseteq\bm{\pi}\in\mathbb{S}_n$, where $\bm{\pi}_i\subseteq\bm{\pi}_u$ are pairwise disjoint partitions of $\bm{\pi}_u$. Note that each entry $\pi_u(j)$ of $\bm{\pi}_u$ is a mapping from $[n]$ to $[n]$.
\end{definition}
The input unit a) is $\mathbb{S}_n$-invariant if $f_u$ is a user-defined $\mathbb{S}_n$-invariant distribution. The sum unit b) is $\mathbb{S}_n$-invariant if all its input units are $\mathbb{S}_n$-invariant. However, it is generally difficult to fulfill the $\mathbb{S}_n$-invariance of the product unit c). Even if the inputs $p_i$ of c) are $\mathbb{S}_n$-invariant, then c) is only \emph{partially} $\mathbb{S}_n$-invariant \citep{papez2024sum}. The partial invariance means that the subgraph $\mathbf{G}_u$ can be permuted only in its individual partitions $\lbrace\mathbf{G}_i\rbrace_{i\in\mathsf{in}(u)}$. \cref{def:conditions-sn-invariance}(c) is stricter in this sense, as it requires that $\mathbf{G}_u$ can also be permuted among its partitions (i.e., permuting the nodes and the corresponding edges between $\mathbf{G}_v$ and $\mathbf{G}_w$ for $v,w\in\mathsf{in}(u)$).

To find an inherently $\mathbb{S}_n$-invariant PGC, we consider that PCs are discrete latent variable models where each sum unit induces a categorical latent variable over its inputs \citep{peharz2016latent,zhao2016unified,trapp2019bayesian}. The consequence of \cref{def:conditions-sn-invariance} is that the latent representation of a PGC must be $\mathbb{S}_n$-invariant for all its components.
\begin{proposition}{(Latent representation of inherently $\mathbb{S}_n$-invariant PGCs.)}\label{prop:latent-representation-sn-invariance}
    A tractable PGC (\cref{prop:tractability-pgc}) in its latent representation,
    \begin{equation}\label{eq:latent-representation}
        p(\mathbf{G})
        =
        \hspace{-4pt}\sum_{\mathbf{z}\in\mathsf{dom}(\mathbf{Z})}\hspace{-4pt}p(\mathbf{G}|\mathbf{z})p(\mathbf{z})
        ,
    \end{equation}
    is inherently $\mathbb{S}_n$-invariant if $p(\bm{\pi}\mathbf{G}|\mathbf{z})=p(\mathbf{G}|\mathbf{z})$ for all $\bm{\pi}\in\mathbb{S}_n$ and all $\mathbf{z}\in\mathsf{dom}(\mathbf{Z})$, where $\mathbf{Z}$ are discrete latent variables of all sum units.
\end{proposition}

One example of an inherently $\mathbb{S}_n$-invariant PGC---which satisfies \cref{prop:latent-representation-sn-invariance}---can be obtained by assuming that the slices of $\mathbf{X}$ and $\mathbf{A}$ are conditionally independent and identically distributed (i.i.d.), as formulated in \cref{prop:factorized-pgcs}.

\begin{proposition}{(Inherently $\mathbb{S}_n$-invariant PGCs through the conditional i.i.d.\ assumption.)}\label{prop:factorized-pgcs}
    A tractable PGC (\cref{prop:tractability-pgc}) is inherently $\mathbb{S}_n$-invariant if its components are factorized as
    \begin{equation}\label{eq:iid-components}
        p(\mathbf{G}^n|\mathbf{z},n)
        =
        \prod_{i\in[n]}p(\mathbf{X}_i|\mathbf{z},n)\prod_{j\in[n]}p(\mathbf{A}_{ij}|\mathbf{z},n)
        .
    \end{equation}
    % where the parameters $\theta^X_{\mathbf{z}}$ of $p(\mathbf{X}_i|\mathbf{z},n)$ are the same for each $i\in[n]$ but distinct for all $\mathbf{z}\in\mathsf{dom}(\mathbf{Z})$, and, similarly, the parameters $\theta^A_{\mathbf{z}}$ of $p(\mathbf{A}_{ij}|\mathbf{z},n)$ are the same for each $i,j\in[n]$ but distinct for all $\mathbf{z}\in\mathsf{dom}(\mathbf{Z})$.
\end{proposition}

Note that the PGC from \cref{prop:factorized-pgcs} captures the correlations both between and within the nodes and edges through the sum in \eqref{eq:latent-representation}, i.e., by conditioning on possibly exponentially many $\mathbf{z}\in\mathsf{dom}(\mathbf{Z})$. However, the i.i.d.\ assumption means that the product terms in \eqref{eq:iid-components} share the parameters for all the slices of $\mathbf{X}$ and, separately, for all the slices of $\mathbf{A}$. Naturally, this sharing mechanism yields less expressive power than having distinct parameters for all the slices. We visualize this PGC in \cref{sec:proofs} (\cref{fig:inherently-sn-invariant-pgc}). Note also that this shared parameterization must be unique for each component $p(\mathbf{G}^n|\mathbf{z},n)$, i.e., for each $\mathbf{z}\in\mathsf{dom}(\mathbf{Z})$; otherwise, the expressive power would drop even further by merging duplicate components of \eqref{eq:latent-representation} into a single one. This strict construction prevents adopting most of the recent design patterns that share the parameters across the components \citep{peharz2020einsum,loconte2024subtractive,loconte2024relationship}. Therefore, we further consider strategies to ensure $\mathbb{S}_n$-invariance (\cref{def:sn-graph-invariance}) even with these traditional $\mathbb{S}_n$-sensitive PCs, though at the cost of not fulfilling \cref{def:tractability-sn-invariance}, as explained next.

\vspace{-4pt}
\subsection{$\mathbb{S}_n$-invariance by marginalization}\label{sec:sn-invariance-by-marginalization}
\vspace{-2pt}
The general way to achieve $\mathbb{S}_n$-invariance is to augment $p(\mathbf{G}^n|n)$ by a latent random variable, $\bm{\pi}\in\mathbb{S}_n$, and compute the marginal probability distribution,
\begin{equation}\label{eq:exact}
    p(\mathbf{G}^n|n)\coloneqq\sum_{\bm{\pi}\in\mathbb{S}_n}p(\mathbf{G}^n|\bm{\pi},n)p(\bm{\pi}|n)
    ,
\end{equation}
which takes $\mathbf{G}^n$ and evaluates the same distribution, $p(\cdot|\bm{\pi},n)$, for all orderings $\bm{\pi}\in\mathbb{S}_n$ of $\mathbf{G}^n$. The marginalization ensures that \eqref{eq:exact} satisfies the standard definition of $\mathbb{S}_n$-invariance (\cref{def:sn-graph-invariance}) for arbitrary permutation-sensitive $p(\cdot|\bm{\pi},n)$ and uniform $p(\bm{\pi}|n)$. Now, consider that $p(\cdot|\bm{\pi},n)$ follows Assumptions \ref{ass:smoothness}-\ref{ass:tractable-input-units}, and that the summation in \eqref{eq:exact} is just a sum unit. Then, \eqref{eq:exact} yields exact values and thus complies with \cref{def:tractability-pc}(A). However, the factorial complexity of the sum violates the polynomial requirement of \cref{def:tractability-pc}(B). For this reason, this approach does not adhere to the tractable $\mathbb{S}_n$-invariance (\cref{def:tractability-sn-invariance}).
% Indeed, \eqref{eq:exact} is computationally infeasible for all but very small graphs. For example, even for $\mathbf{G}^n$ representing a molecule with, say, $n=9$ atoms (the QM9 dataset \citep{ramakrishnan2014quantum}), \eqref{eq:exact} requires 362880 evaluations of $p(\cdot|\bm{\pi},n)$.

\textbf{Asymmetry implies intractability.} One possible way to tame the factorial complexity of \eqref{eq:exact} is to simply reduce the number of sum terms \citep{wagstaff2022universal}. This idea still satisfies \cref{def:tractability-pc}(B)---without violating Assumptions \ref{ass:smoothness}-\ref{ass:tractable-input-units}---however, we immediately loose \cref{def:tractability-pc}(A). The reason is that breaking the symmetry by neglecting even a single ordering, $\bm{\pi}\in\mathbb{S}_n$, makes \eqref{eq:exact} inexact. To see this, let $\widehat{\mathbb{S}}_n\subset\mathbb{S}_n$ be a subset consisting of $n!-1$ orderings, then it holds that $p(\mathbf{G}^n|n)\geq\sum_{\bm{\pi}\in\widehat{\mathbb{S}}_n}p(\mathbf{G}^n,\bm{\pi}|n)$.

\vspace{-4pt}
\subsection{$\mathbb{S}_n$-invariance by sorting}\label{sec:sn-invariance-by-sorting}
\vspace{-2pt}
To make PGCs $\mathbb{S}_n$-invariant, we can simply sort $\mathbf{G}$ into a pre-defined canonical ordering, $\bm{\pi}_c$, each time before it enters the model as input. The model then returns the same value for each permutation of $\mathbf{G}$ and is therefore $\mathbb{S}_n$-invariant. However, this design choice does not satisfy \cref{def:tractability-sn-invariance} since we no longer operate with the exact likelihood \eqref{eq:exact} but only its lower bound. To demonstrate this, consider the variational evidence lower bound on the logarithm of \eqref{eq:exact},
\begin{equation}\label{eq:elbo-generic}
    \log p(\mathbf{G}^n|n)\!\geq\!\mathbb{E}_{q(\bm{\pi}|\mathbf{G}^n)}\!\left[\log p(\mathbf{G},\bm{\pi}|n)\right] \!+\! \mathcal{H}(q(\bm{\pi}|\mathbf{G}^n))
    ,
\end{equation}
where $q(\bm{\pi}|\mathbf{G}^n)$ is the variational posterior distribution, and $\mathcal{H}$ is the differential entropy. If there is only a single canonical ordering, $\bm{\pi}_c$, then $q(\bm{\pi}|\mathbf{G}^n)$ concentrates all its mass to a single point, and \eqref{eq:elbo-generic} becomes $\log p(\mathbf{G}^n|n)\geq\log p(\mathbf{G}^n|\bm{\pi}_c,n) + c$ where $c=\log p(\bm{\pi}_c|n)$. This approach significantly reduces the computational complexity of \eqref{eq:exact} from $\mathcal{O}(n!)$ to the complexity of a selected sorting algorithm. Here, we can see that satisfying \cref{def:tractability-sn-invariance} is uneasy. On the one hand, we reduce the factorial complexity down to the polynomial one (satisfying \cref{def:tractability-pc}(B)). On the other hand, we pay for this by involving variational approximation (sacrificing \cref{def:tractability-pc}(A)).

\textbf{Some orderings are more suitable than others.} Which of the $n!$ orderings to choose? Not all orderings are suitable. Random ordering can harm the performance of the model. However, a subset of $\mathbb{S}_n$ leads to a good performance. We refer to this collection as canonical orderings. Some rely on specific domain knowledge, whereas others are completely domain agnostic. \cref{sec:orderings} (\cref{fig:orderings}) shows an unnormalized empirical distribution over adjacency matrices for a random ordering and four canonical orderings. The ordered adjacency matrices are more structured and highly concentrated near the diagonal, which simplifies the learning of $p(\mathbf{G})$ and increases the chance of sampling semantically valid graphs (Figures~\ref{fig:ppgc-orderings-qm9}~and~\ref{fig:ppgc-orderings-zinc250k}).

\textbf{PGCs for undirected acyclic graphs.} The structural properties of $\mathbf{G}$ are characterized by the layout of $\mathbf{A}$\footnote{For example, if $\mathbf{A}$ is asymmetric with non-zero diagonal entries, then $\mathbf{G}$ is a directed cyclic graph.}. We consider PGCs for undirected acyclic graphs, where $\mathbf{A}$ is uniquely determined by $N\tfrac{N-1}{2}n_A$ entries in its lower triangular part, $\mathbf{L}$, which allows us to avoid the repeated entries and improve the scalability. Consequently, $p(\mathbf{G}^n|n)$ in \eqref{eq:pgc} is the following joint probability distribution:
\begin{equation}\nonumber
    p(\mathbf{G}^n|n)
    \coloneqq
    p(\mathbf{X}^n,\mathbf{L}^n|n)
    % \coloneqq
    % \sum_{i\in\mathsf{in}(u)}w_ip_i(\mathbf{X}^n|n)p_i(\mathbf{L}^n|n)
    .
\end{equation}
This PGC can be architectured in many different ways. In this paper, we focus on splitting $p(\mathbf{X}^n,\mathbf{L}^n|n)$ into two main parts, as illustrated in \cref{fig:pgc}. The first models the nodes, taking $\mathbf{X}^n$ as input, whereas the second models the edges, taking row-flattened $\mathbf{L}^n$ as input. The last layer of both these parts is the sum layer with $n_c$ nodes. To capture the correlations between the nodes and edges, we connect the outputs of these two parts by the product layer with $n_c$ units and then aggregate its outputs by the sum layer with a single unit. The higher the values of $n_c$, the better we capture the interactions between the nodes and edges. In \cref{sec:experiments}, we compare different node-PC and edge-PC parts implementations.

\begin{table*}[!t]
    \renewcommand{\arraystretch}{0.8}
    \caption{\emph{Unconditional generation on the QM9 and Zinc250k datasets}. The mean value and standard deviation of the molecular metrics for the baseline intractable DGMs (top) and various implementations of the $\bm{\pi}$PGCs that rely on the sorting to ensure the $\mathbb{S}_n$-invariance (bottom). The results are computed over five runs with different initialization. The \textcolor{c1}{1st}, \textcolor{c2}{2nd}, and \textcolor{c3}{3rd} best results are highlighted in colors.}
    \vspace{-15pt}
    \label{tab:generation-unco}
    \begin{center}
        \begin{scriptsize}
            \begin{tabular}{lllllllllll}
\toprule
 & \multicolumn{5}{c}{QM9} & \multicolumn{5}{c}{Zinc250k} \\
\cmidrule(lr){2-6}
\cmidrule(lr){7-11}
Model & Valid$\uparrow$ & NSPDK$\downarrow$ & FCD$\downarrow$ & Unique$\uparrow$ & Novel$\uparrow$ & Valid$\uparrow$ & NSPDK$\downarrow$ & FCD$\downarrow$ & Unique$\uparrow$ & Novel$\uparrow$ \\
\midrule
GraphAF & 74.43$\pm$2.55 & 0.021$\pm$0.003 &  \phantom{0}5.27$\pm$0.40 & 88.64$\pm$2.37 & 86.59$\pm$1.95 & 68.47$\pm$0.99 & 0.044$\pm$0.005 & 16.02$\pm$0.48 & \phantom{0}98.64$\pm$0.69 & 100.00$\pm$0.00 \\
GraphDF & \color{c3} \textbf{93.88$\pm$4.76} & 0.064$\pm$0.000 & 10.93$\pm$0.04 & 98.58$\pm$0.25 & \color{c1} \textbf{98.54$\pm$0.48} & \color{c3} \textbf{90.61$\pm$4.30} & 0.177$\pm$0.001 & 33.55$\pm$0.16 & \phantom{0}99.63$\pm$0.01 &  \phantom{0}99.99$\pm$0.01 \\
MoFlow & 91.36$\pm$1.23 & 0.017$\pm$0.003 &  \phantom{0}4.47$\pm$0.60 & 98.65$\pm$0.57 & 94.72$\pm$0.77 & 63.11$\pm$5.17 & 0.046$\pm$0.002 & 20.93$\pm$0.18 & \phantom{0}99.99$\pm$0.01 & 100.00$\pm$0.00 \\
EDP-GNN & 47.52$\pm$3.60 & 0.005$\pm$0.001 &  \phantom{0}2.68$\pm$0.22 & 99.25$\pm$0.05 & 86.58$\pm$1.85 & 82.97$\pm$2.73 & 0.049$\pm$0.006 & 16.74$\pm$1.30 & \phantom{0}99.79$\pm$0.08 & 100.00$\pm$0.00 \\
GraphEBM &  \phantom{0}8.22$\pm$2.24 & 0.030$\pm$0.004 &  \phantom{0}6.14$\pm$0.41 & 97.90$\pm$0.14 & \color{c3} \textbf{97.01$\pm$0.17} &  \phantom{0}5.29$\pm$3.83 & 0.212$\pm$0.005 & 35.47$\pm$5.33 & \phantom{0}98.79$\pm$0.15 & 100.00$\pm$0.00 \\
SPECTRE & 87.30$\pm$n/a & 0.163$\pm$n/a & 47.96$\pm$n/a & 35.70$\pm$n/a & \color{c2} \textbf{97.28$\pm$n/a} & 90.20$\pm$n/a & 0.109$\pm$n/a & 18.44$\pm$n/a & \phantom{0}67.05$\pm$n/a & 100.00$\pm$n/a \\
GDSS & \color{c2} \textbf{95.72$\pm$1.94} & \color{c3} \textbf{0.003$\pm$0.000} &  \phantom{0}2.90$\pm$0.28 & 98.46$\pm$0.61 & 86.27$\pm$2.29 & \color{c1} \textbf{97.01$\pm$0.77} & \color{c1} \textbf{0.019$\pm$0.001} & 14.66$\pm$0.68 & \phantom{0}99.64$\pm$0.13 & 100.00$\pm$0.00 \\
DiGress & \color{c1} \textbf{99.00$\pm$0.10} & 0.005$\pm$n/a & \color{c1} \phantom{0}\textbf{0.36$\pm$n/a} & 96.20$\pm$n/a & 33.40$\pm$n/a & \color{c2} \textbf{91.02$\pm$n/a} & 0.082$\pm$n/a & 23.06$\pm$n/a & \phantom{0}81.23$\pm$n/a & 100.00$\pm$n/a \\
GRAPHARM & 90.25$\pm$n/a & \color{c2} \textbf{0.002$\pm$n/a} & \color{c3} \phantom{0}\textbf{1.22$\pm$n/a} & 95.62$\pm$n/a & 70.39$\pm$n/a & 88.23$\pm$n/a & 0.055$\pm$n/a & 16.26$\pm$n/a & \phantom{0}99.46$\pm$n/a & 100.00$\pm$n/a \\
\midrule
BT & 78.15$\pm$0.70 & 0.004$\pm$0.001 & \phantom{0}1.68$\pm$0.13 & \color{c2} \textbf{99.66$\pm$0.10} & 93.20$\pm$0.36 & 17.00$\pm$0.55 & 0.050$\pm$0.002 & \color{c3} \textbf{\phantom{0}9.42$\pm$0.36} & 100.00$\pm$0.00 & 100.00$\pm$0.00 \\
LT & 62.81$\pm$2.69 & 0.007$\pm$0.001 & \phantom{0}2.57$\pm$0.24 & \color{c1} \textbf{99.72$\pm$0.05} & 95.95$\pm$0.35 & \phantom{0}4.11$\pm$0.24 & 0.056$\pm$0.001 & 11.53$\pm$0.19 & 100.00$\pm$0.00 & 100.00$\pm$0.00 \\
RT & 81.83$\pm$1.62 & 0.003$\pm$0.000 & \phantom{0}1.29$\pm$0.04 & 99.37$\pm$0.07 & 90.67$\pm$0.40 & \phantom{0}6.90$\pm$0.58 & 0.047$\pm$0.000 & \phantom{0}9.59$\pm$0.39 & \color{c3} \textbf{100.00$\pm$0.00} & \color{c3} \textbf{100.00$\pm$0.00} \\
RT-S & 88.83$\pm$0.75 & \color{c1} \textbf{0.002$\pm$0.000} & \color{c2} \textbf{\phantom{0}1.11$\pm$0.01} & 99.38$\pm$0.06 & 88.49$\pm$0.45 & 14.66$\pm$0.66 & \color{c3} \textbf{0.043$\pm$0.002} & \color{c1} \textbf{\phantom{0}8.78$\pm$0.34} & \color{c2} \textbf{100.00$\pm$0.00} & \color{c2} \textbf{100.00$\pm$0.00} \\
HCLT & 89.00$\pm$0.17 & 0.003$\pm$0.000 & \phantom{0}1.45$\pm$0.09 & \color{c3} \textbf{99.45$\pm$0.04} & 90.62$\pm$0.43 & 23.67$\pm$0.45 & \color{c2} \textbf{0.035$\pm$0.000} & \color{c2} \textbf{\phantom{0}8.93$\pm$0.06} & \color{c1} \textbf{100.00$\pm$0.00} & \color{c1} \textbf{100.00$\pm$0.00} \\
\bottomrule
\end{tabular}
        \end{scriptsize}
    \end{center}
    \vspace{-15pt}
\end{table*}

\vspace{-4pt}
\section{Experiments}\label{sec:experiments}
\vspace{-2pt}
We evaluate PGCs in the following three experiments. First, we compare $\mathbb{S}_n$-sensitive and $\mathbb{S}_n$-invariant PGCs. Second, we assess their ability to generate novel and semantically valid graphs. Third, to demonstrate their aptness for tractable probabilistic inference, we show the generation of new graphs conditionally on known graph substructures. The experiments are performed in the context of the computational design of molecular graphs. We provide the code at \url{https://github.com/mlnpapez/PGC}.

\textbf{Molecular design.} DGMs for molecular graphs are important in discovering drugs and materials with desired properties. Given a dataset of molecular graphs, the task is to learn a probability distribution of chemically valid molecules, $p(\mathbf{G})$. This is a complex combinatorial task, as not all combinations of atoms and bonds can be connected, but the connections must adhere to chemical valency rules. The aim is that $p(\mathbf{G})$ can produce molecules that were unseen during the training but satisfy the valency constraints.

\textbf{Datasets.} We test the performance of PGCs on QM9 \citep{ramakrishnan2014quantum} and ZINC250k \citep{irwin2012zinc} datasets, which are two standard benchmarks that are often used to assess DGMs for molecular design. We summarize the statistics of these datasets in \cref{tab:dataset-statistics} and describe their preprocessing in \cref{sec:datasets}.

\textbf{Metrics.} We evaluate \emph{valid}, \emph{unique}, and \emph{novel} molecules as the percentage of all generated molecules that satisfy chemical valency rules (without resorting to any corrections mechanisms), the percentage of valid molecules that are not a duplicate of some other generated molecule, and the percentage of valid and unique molecules that are not in the training data, respectively. We further compute Fr\'{e}chet ChemNet distance (FCD) \citep{preuer2018frechet}, which is the distance between the generated and training molecules computed from the activations of the second last layer of ChemNet, and neighborhood subgraph pairwise distance kernel (NSPDK) MMD \citep{costa2010fast}, which is the distance between the generated and test molecules such that both the node and edge features are accounted for.

\textbf{PGC variants.} We distinguish PGCs with the following $\mathbb{S}_n$-invariance principles: inherent $\mathbb{S}_n$-invariance (\cref{sec:sn-invariance-inherent}), $\mathbb{S}_n$-invariance by marginalization (\cref{sec:sn-invariance-by-marginalization}), $\mathbb{S}_n$-invariance by conditioning on a canonical ordering, $p(\mathbf{G}|\bm{\pi}_c)$, (\cref{sec:sn-invariance-by-sorting}), and no $\mathbb{S}_n$-invariance, which we refer to as $i$PGC, $n!$PGC, $\bm{\pi}$PGC, and $s$PGC, respectively. All these models follow the design template in \cref{fig:pgc}.

\begin{figure}[!t]
    \begin{center}
        \centerline{\input{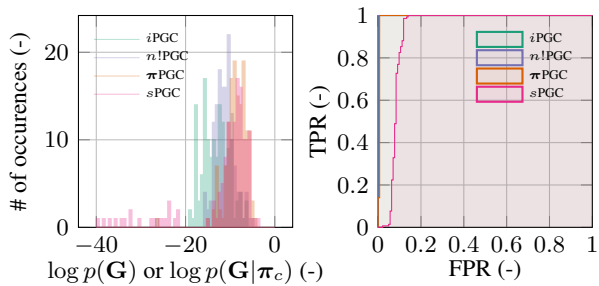}}
        \vspace{-7pt}
        \caption{\emph{$\mathbb{S}_n$-sensitivity of PGCs.} The histograms of $\log p(\mathbf{G})$ and $\log p(\mathbf{G}|\bm{\pi}_c)$ (left) and the AUC (right) for different $\mathbb{S}_n$-invariance mechanisms of PGCs in the context of anomaly detection.}
        \vspace{-35pt}
        \label{fig:sn-sensitivity}
    \end{center}
\end{figure}

\textbf{$\mathbb{S}_n$-sensitivity of PGCs.} A properly designed graph DGM should be able to recognize a single instance of a graph, $\mathbf{g}^n$, up to all its $n!$ permutations. This is especially important in anomaly detection since we do not want our model to detect different graph permutations as anomalous. Therefore, we compare the $\mathbb{S}_n$-invariant PGCs from \cref{sec:pgcs} with an $\mathbb{S}_n$-sensitive PGC on a synthetic anomaly detection example. To carry out the experiment in a feasible time span (due to the $n!$ complexity of \eqref{eq:exact}), we restrict ourselves to only a subset of the QM9 dataset containing molecules with up to 6 atoms. We train the models on molecules with up to 5 atoms (in-distribution data) and use the molecules with 6 atoms as the anomalies (out-of-distribution data), making these two subsets perfectly balanced in their size. However, we randomly permute 20\% of the in-distribution molecules during the evaluation. \cref{fig:sn-sensitivity} draws the histogram of the exact log-likelihood $\log p(\mathbf{G})$ or its lower bound $\log p(\mathbf{G}|\bm{\pi}_c)$, and the area under the curve (AUC) obtained by computing the true positive rate (TPR) and the true negative rate (TNR). All the $\mathbb{S}_n$-invariant PGCs correctly recognize the permuted molecules as in-distribution data. In contrast, the $\mathbb{S}_n$-sensitive PGC incorrectly recognizes them as out-of-distribution data by assigning them very low log-likelihoods. This behavior has a severe impact on the AUC metric. While the $\mathbb{S}_n$-invariant PGCs achieve near-perfect AUC, the $\mathbb{S}_n$-sensitive PGC suffers from non-zero~FPR.

In this experiment, the number of parameters in $i$PGC is 10x higher than in $\bm{\pi}$PGC, $n!$PGC, or $s$PGC. Still, we observe in \cref{fig:sn-sensitivity} that $\bm{\pi}$PGC consistently outperforms $i$PGC, confirming our previous claim about the reduced expressive power of $i$PGCs (\cref{sec:pgcs}). Therefore, we continue our experiments only with $\bm{\pi}$PGC, leaving $n!$PGC and $s$PGC due to their factorial complexity and $\mathbb{S}_n$-sensitivity, respectively.

\textbf{$\bm{\pi}$PGC variants.} The PGCs (\cref{sec:pgcs}) can be instantiated in different ways, depending on the region graph (RG) \citep{dennis2012learning} that is used to build the architecture of the node-PC and edge-PC (\cref{fig:pgc}). We consider the following types of region graphs: binary tree (BT) \citep{loconte2024subtractive}, linear tree (LT) \citep{loconte2024relationship}, randomized tree (RT) \citep{peharz2020einsum,peharz2020random}, randomized tree with synchronized permutations between the node and edge PCs (RT-S), and the hidden Chow-Liu tree (HCLT) \citep{liu2021tractable}. Note that while the first four variants do not rely on data to build the RG, the last one learns the structure of the Chow-Liu tree from data. We consider that all these RGs are used to build a tensorized, monotonic architecture (\cref{def:tensorized-pcs}). We perform an extensive gridsearch over the hyper-parameters of these architectures, as detailed in \cref{sec:experimental-details}. The ordering is also a hyper-parameter and covers all cases discussed in \cref{sec:orderings}.

\textbf{Baselines.} We compare $\bm{\pi}$PGCs with the following intractable DGMs: GraphAF \citep{shi2020graphaf}, GraphDF \citep{luo2021graphdf}, MoFlow \citep{zang2020moflow}, EDP-GNN \citep{niu2020permutation}, GraphEBM \citep{liu2021graphebm}, SPECTRE \citep{martinkus2022spectre}, GDSS \citep{jo2022score}, DiGress \citep{vignac2023digress}, GraphARM \citep{kong2023autoregressive}. Additional details about these models are given in \cref{sec:baselines}.

\textbf{Unconditional molecule generation.} \cref{tab:generation-unco} shows that all the $\bm{\pi}$PGCs variants deliver a competitive performance across all metrics on the QM9 dataset. Most notably, the RT-S variant scores \textcolor{c1}{1st} in NSPDK and \textcolor{c2}{2nd} in FCD. For the Zinc250k dataset, the RT-S model is \textcolor{c3}{3rd} in NSPDK and \textcolor{c1}{1st} in FCD, whereas the HCLT model is \textcolor{c2}{2nd} in both these metrics. However, the validity drops for the Zinc250k dataset, where even the best $\bm{\pi}$PGC variant (HCLT) delivers only 23\% validity. Recall that we do not use any validity corrections mechanism \citep{zang2020moflow} to satisfy the valency rules but rely on our model to learn them on~its~own. The validity metric is considered unreliable, as it can be artificially increased by generating simpler molecules even with rule-based systems \citep{preuer2018frechet}. The NSPDK and FCD metrics are more robust, revealing better diversity of the generated molecules and similarity to real molecules.

\begin{figure}[!t]
    \centering
    \includesvg[width=0.9\linewidth]{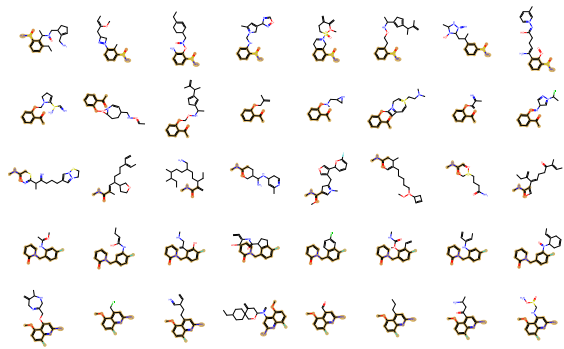}
    \vspace{-7pt}
    \caption{\emph{Conditional generation on the Zinc250k dataset.} The \textcolor{c6}{yellow} area highlights the known part of the molecule. There is one such known part per row. Each column corresponds to a new molecule generated conditionally on the known part.}
    \vspace{-12pt}
    \label{fig:cond-zinc250k-ptree}
\end{figure}

\textbf{Conditional molecule generation.} To illustrate that PGCs can answer probabilistic inference queries, \cref{fig:cond-zinc250k-ptree} displays conditional molecule generation for the RT-S variant, where we can see that each molecule's newly generated part varies in size and composition. More examples of the conditional generation, along with the corresponding molecular metrics, are provided in \cref{sec:additional-results}, which also contains a detailed comparison of all $\bm{\pi}$PGC variants for different orderings in the context of the unconditional generation.

\vspace{-4pt}
\section{Conclusion and Limitations}
\vspace{-2pt}
\textbf{Conclusion.} We have proposed PGCs---tractable probabilistic models for graphs---and specified the conditions for their tractability through the extended definition of $\mathbb{S}_n$-invariance. Furthermore, we have developed marginalization padding, which conveniently utilizes the tractability of PGCs to handle the variable-size nature of graphs. We have shown that designing PGCs with the inherently build-in, tractable $\mathbb{S}_n$-invariance---i.e., without using any specific $\mathbb{S}_n$-invariance mechanisms---requires rather strict conditional i.i.d.\ assumptions on modeling the nodes and edges, which undermines the expressive power of such PGCs. Therefore, we have introduced PGCs that are $\mathbb{S}_n$-invariant through the sorting. Despite not complying with the strict form of tractable $\mathbb{S}_n$-invariance, they answer the same inference queries as conventional PCs. Importantly, these PGCs outperform, or are competitive with, existing intractable DGMs in most standard molecular metrics except for the validity on the Zinc250k dataset. %This result comes even at the cost of not working with the exact likelihood but only its lower bound.

\textbf{Limitations.} The low validity on the Zinc250k dataset can easily be improved by rejection sampling, which is computationally cheap with PGCs, as they produce samples in a single pass through the network. This contrasts with autoregressive and diffusion models, where high validity is substantially more important since they generate a single sample iteratively, making several expensive passes through the model. We conjecture that the low validity comes from the template in \cref{fig:pgc}, where the node and edge PCs are connected only through a mixture of independent components. This architecture captures the correlations between the nodes and edges by a sufficiently high number of components. However, it can be improved by connecting the node and edge PCs also between their lower layers. To make this enhancement, we would require PCs with an input layer that allows for hybrid variables (as there are different numbers of categories for the nodes and edges), which is currently unavailable with the existing PC libraries. We will address this limitation in future work.

\begin{acknowledgements}
The authors acknowledge the support of the GA\v{C}R grant no.\ GA22-32620S and the OP VVV funded project CZ.02.1.01/0.0/0.0/16\_019/0000765 ``Research Center for Informatics''.
\end{acknowledgements}

\balance
\bibliography{uai2025}

\newpage

\onecolumn

\title{Probabilistic Graph Circuits: Deep Generative Models for Tractable\\Probabilistic Inference over Graphs\\(Supplementary Material)}
\maketitle

\appendix
\section{Permutation Invariance}

\textbf{$\mathbb{S}_n$-invariance.} Exchangeable data structures, including sets, graphs, partitions, and arrays \citep{orbanz2014bayesian}, have a factorial number of possible configurations (orderings). Permutation invariance says that no matter a selected configuration, the probability of an exchangeable data structure has to remain the same.

To define the permutation invariance of a probability distribution, let $\mathbb{S}_n$ be a finite symmetric group of a set of $n$ elements. This is a set of all $n!$ permutations of $[n]\coloneqq(1,\ldots,n)$, where any of its members, $\bm{\pi}\in\mathbb{S}_n$, permutes an $n$-dimensional vector, $\mathbf{X}\coloneqq(X_1,\ldots,X_n)$, as follows: $\bm{\pi}\mathbf{X}=(X_{\pi(1)},\ldots,X_{\pi(n)})$. % The probability distribution $p$ is permutation invariant iff $p(\mathbf{X})=p(\bm{\pi}\mathbf{X})$ for all $\bm{\pi}\in\mathbb{S}_n$.

\begin{definition}{($\mathbb{S}_n$-invariance of vectors).}\label{def:sn-invariance}
    The probability distribution $p$ is permutation invariant iff $p(\mathbf{X})=p(\bm{\pi}\mathbf{X})$ for all $\bm{\pi}\in\mathbb{S}_n$. $\mathbf{X}$ is permutation invariant if $p(\mathbf{X})$~is.
\end{definition}

\section{Circuits}
\label{sec:circuits}

In this section, to simplify the notation, we consider that the size of $\mathbf{X}$ is fixed, $N=n$. Therefore, we leave the explicit dependence on $n$ out of $p(\mathbf{X}^n|n)$.

\begin{definition}[Circuit \citep{vergari2021compositional}]\label{def:circuit}
    A \emph{circuit} $c$ over random variables $\mathbf{X}\coloneqq\lbrace X_1,\ldots,X_n\rbrace$ is a parameterized, directed, acyclic computational graph encoding a function $c(\mathbf{X})$. It contains three types of computational \emph{units}: \emph{input}, \emph{product}, and \emph{sum}. All sum and product units receive the outputs of other units as inputs. We denote the set of inputs of a unit $u$ as $\mathsf{in}(u)$. Each unit $u$ encodes a function $c_u$ over a subset of the random variables, $\mathbf{X}_u\subseteq\mathbf{X}$, which we refer to as \emph{scope}. The input unit $c_u(\mathbf{X}_u)\coloneqq f_u(\mathbf{X}_u)$ computes a user-defined parameterized function, $f_u$; the sum unit computes the weighted sum of its inputs, $c_u(\mathbf{X}_u)\coloneqq\sum_{i\in\mathsf{in}(u)}w_ic_i(\mathbf{X}_i)$, where $w_i\in\mathbb{R}$ are the weight parameters; and the product unit computes the product of its inputs, $c_u(\mathbf{X}_u)\coloneqq\prod_{i\in\mathsf{in}(u)}c_i(\mathbf{X}_i)$. The scope of any sum or product unit is the union of its input scopes, $\mathbf{X}_u=\bigcup_{i\in\mathsf{in}(u)}\mathbf{X}_i$. A circuit can have one or multiple root units. The scope of a root unit is $\mathbf{X}$.
\end{definition}

\begin{definition}[Probabilistic circuit]\label{def:probabilistic-circuit}
    A \emph{probabilistic circuit} (PC) over random variables $\mathbf{X}$ is a circuit (\cref{def:circuit}) $c$, such that $\forall \mathbf{x}\in\mathsf{dom}(\mathbf{X}): c(\mathbf{x})\geq 0$, i.e., it is a \emph{non-negative} function for all values of $\mathbf{X}$.
\end{definition}

A PC (\cref{def:probabilistic-circuit}) can encode a possibly unnormalized probability distribution, $p(\mathbf{x})\propto c(\mathbf{x})$.

\textbf{Tractable probabilistic inference.} PCs are \emph{tractable} if they provide \emph{exact} and \emph{efficient} answers to inference queries over arbitrary subsets of $\mathbf{X}$ \citep{choi2020probabilistic,vergari2021compositional}. Here, exact means that the answers do not involve any approximations, and efficient means that the answers are computed in polytime. An arbitrary inference query can be expressed in terms of the expectation $\mathbb{E}_{p(\mathbf{X})}[h(\mathbf{X})]=\int h(\mathbf{X})p(\mathbf{X})d\mathbf{X}$, where $h$ is a function that allows us to formulate a desired query over $\mathbf{X}$ or its part(s). However, to make the computation of $\mathbb{E}_{p}[h]$ tractable, both $p$ and $h$ have to satisfy certain assumptions.

\begin{assumption}[Smoothness \citep{darwiche2002knowledge}]\label{ass:smoothness}
    A circuit $c$ is \emph{smooth} if the inputs of each sum unit $u\in c$ have the same scope, $\forall a,b\in\mathsf{in}(u):\mathbf{X}_a=\mathbf{X}_b$.
\end{assumption}

\begin{assumption}[Decomposability \citep{darwiche2002knowledge}]\label{ass:decomposability}
    A circuit $c$ is \emph{decomposable} if the inputs of each product unit $u\in c$ have pair-wise disjoint scopes $\forall a,b\in\mathsf{in}(u):\mathbf{X}_a\cap\mathbf{X}_b=\varnothing$.
\end{assumption}

\begin{assumption}[Tractable input units]\label{ass:tractable-input-units}
    A circuit $c$ has \emph{tractable input units} if each input unit $u\in c$ admits an algebraically closed-form solution to its integral, $\int_{\mathsf{dom}(\mathbf{Z}_u)}c_u(\mathbf{y}_u,\mathbf{z}_u)d\mathbf{Z}_u$, for any $\mathbf{X}_u\coloneqq\lbrace\mathbf{Y}_u,\mathbf{Z}_u\rbrace$.
\end{assumption}

To satisfy Assumption \ref{ass:tractable-input-units}, we need each $c_u(\mathbf{X}_u)$ to belong to a tractable family of probability distributions \citep{barndorff1978information}.

\begin{assumption}[Compatibility \citep{vergari2021compositional}]\label{ass:compatibility}
   Two circuits $f$ and $g$ over the same random variables $\mathbf{X}$ are \emph{compatible} if they satisfy Assumptions \ref{ass:smoothness} and \ref{ass:decomposability}, and any two product units $a\in f$ and $b\in g$, such that $\mathbf{X}_a=\mathbf{X}_b$, can be rearranged into pair-wise compatible products that decompose in the same way: $(\mathbf{X}_a=\mathbf{X}_b)\Rightarrow (\mathbf{X}_{a_i}=\mathbf{X}_{b_i}$, $a_i$ and $b_i$ are compatible) for some rearrangement of the inputs of $a$ (resp. $b$) into $a_1$, $a_2$ (resp. $b_1$, $b_2$).
\end{assumption}

\begin{definition}[Tractable expectation]\label{def:tractable-expectation}
    Two circuits $f$ and $g$ over the same random variables $\mathbf{X}$ admit tractable expectation, $\mathbb{E}_{g}[f]\coloneqq\int f(\mathbf{X})g(\mathbf{X})d\mathbf{X}$, if they satisfy Assumption \ref{ass:compatibility}, and the product of any two input units $a\in f$ and $b\in g$, such that $\mathbf{X}_a=\mathbf{X}_b$, can be integrated tractably.
\end{definition}

\begin{definition}[Tensorized PCs]\label{def:tensorized-pcs}
    A tensorized PC \citep{peharz2020einsum,peharz2020random,loconte2024subtractive} is a deep learning model of a probability distribution, $p(\mathbf{X})$, over a \emph{fixed-size} vector, $\mathbf{X}\in\mathcal{X}$. The network contains several layers of computational units (similar to neural networks \citep{vergari2019visualizing}). Each layer is defined over its \emph{scope}, $\mathbf{X}_u\subseteq\mathbf{X}$, i.e., a subset of the input. There are three types of layers, depending on the units they encapsulate: sum layer $\mathsf{L}_\mathsf{S}$, product layer $\mathsf{L}_\mathsf{P}$, and input layer $\mathsf{L}_\mathsf{I}$. The units of \emph{input} layers are user-defined probability distributions, $p_{u,i}(\mathbf{X}_u)$. For $n_I$ units, an input layer computes $p_{u,i}(\mathbf{X}_u)$ for $i\in(1,\ldots,n_I)$ and outputs an $n_I$-dimensional vector of probabilities $\mathbf{l}$. The units of \emph{product} layers are factored distributions, applying conditional independence over a pair-wise disjoint partition of their scope. A product layer receives outputs from $n$ layers, $(\mathbf{l}_1,\ldots,\mathbf{l}_n)$, and computes either an Hadamard product, $\mathbf{l}=\odot^n_{i=1}\mathbf{l}_i$, or Kronecker product, $\mathbf{l}=\otimes^n_{i=1}\mathbf{l}_i$. The units of \emph{sum} layers are mixture distributions. For $n_S$ units, a sum layer receives an $n$-dimensional input, $\mathbf{l}$, from a previous layer and computes $\mathbf{W}\mathbf{l}$, where $\mathbf{W}$ is an $n_S\times n$ matrix of row-normalized weights. The output (layer) of a tensorized PC is typically a sum layer.
\end{definition}

\textbf{PCs can only be partially $\mathbb{S}_n$-invariant.} The study of $\mathbb{S}_n$-invariance of PCs has received only limited attention until recently \citep{papez2024sum}, where the authors prove that even if the input units of a PC are permutation invariant, then the traditional structural restrictions---i.e., smoothness (\cref{ass:smoothness}) and decomposability (\cref{ass:decomposability})---make this PC only partially permutation invariant. For this reason, we consider conventional PCs to be permutation-sensitive distributions in the present paper.

\section{Related Work}
\label{sec:related-work}

\textbf{Intractable probabilistic models.} Practically all canonical types of DGMs have been extended to graph-structured data. These DGMs learn $p(\mathbf{G})$ (or its approximation) in different ways. Autoregressive models rely on the chain rule of probability to construct $\mathbf{G}$ by sequentially adding new nodes and edges based on the previous ones \citep{you2018graphrnn,liao2019efficient}. Variational autoencoders train an encoder and a decoder to map between space of graphs and latent space \citep{simonovsky2018graphvae,liu2018constrained,ma2018constrained,grover2019graphite,kwon2019efficient,samanta2020nevae}. Generative adversarial networks train (i) a generator to map from latent space to space of graphs, and (ii) a discriminator to distinguish whether the graphs are synthetic or real \citep{de2018molgan,bojchevski2018netgan}. Normalizing flows use the change of variables formula to transform a base distribution on latent space to a distribution on space of graphs, $p(\mathbf{G})$, using invertible neural networks \citep{liu2019graph,luo2021graphdf}. Energy-based models define $p(\mathbf{G})$ by the Boltzmann distribution, whose energy function assigns low energies to correct graphs and high energies to incorrect graphs \cite{liu2021graphebm}. Diffusion models systematically noise and denoise trajectories of graphs based on forward and backward diffusion processes, respectively \citep{jo2022score,huang2022graphgdp,vignac2023digress,kong2023autoregressive,hua2024mudiff}. All these models rely on deep neural networks that contain non-linear transformations, which prevents tractable computation of probabilistic inference queries.

\textbf{Tractable probabilistic models.} TPMs for graphs in the variable-size tensor representation considered in this paper---where both the nodes and edges are attributed---have received no attention. Existing models are designed either for general (possibly cyclic) graphs without edge attributes \citep{zheng2018learning,errica2024tractable} or for graphs of specific relational structures \citep{nath2015learning,loconte2023turn,papez2024sum}, including trees (typically also without edge attributes) and knowledge graphs. These TPMs do not encode a fully permutation invariant probability distribution over a graph, with the only exceptions \citep{nath2015learning,papez2024sum}, which are permutation invariant only for certain graph sub-structures. Additionally, certain TPMs do not encode a probability distribution over a whole graph, but only its triplet relations \citep{loconte2023turn}, thus proving only local tractable inference. Moreover, adopting the pseudo-likelihood in the GSPN model \citep{errica2024tractable} renders the inference intractable even from the standard perspective of tractability (\cref{def:tractability-pc}). Aside from \citep{papez2024sum}, these TPMs do not model the cardinality of a graph, which hinders their generative capabilities. Similarly, excluding \citep{loconte2023turn} and partially \citep{nath2015learning,papez2024sum}, these TPMs rely on a fixed graph ordering.

\section{Tractable Inference Queries Over Graphs}
\label{sec:queries}
Probabilistic inference queries over graphs---such as marginalization, conditioning, and expectation---are important in various applications. The examples include explainable anomaly detection (marginalization \cite{ying2019gnnexplainer}), retro-synthesis (conditioning \cite{igashov2024retrobridge}), and uncertainty quantification (expectation \cite{yang2023explainable}). However, many graph DGMs are intractable and thus do not permit even the most basic inference queries. On the contrary, PGCs are tractable DGMs that allow for a broad range of inference queries.

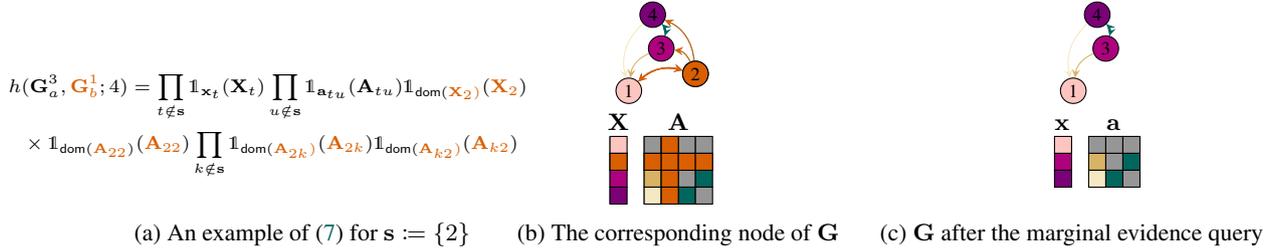
\begin{figure*}[ht]
    \begin{center}
        \centerline{\newcommand\grid{8pt}

\definecolor{c0}{RGB}{243,246,249}
\definecolor{c1}{RGB}{27,158,119}
\definecolor{c2}{RGB}{117,112,179}
\definecolor{c3}{RGB}{217,95,2}
\definecolor{c4}{RGB}{231,41,138}
\definecolor{c5}{RGB}{150,150,150}

\definecolor{n1}{RGB}{252,197,192}
\definecolor{n2}{RGB}{247,104,161}
\definecolor{n3}{RGB}{174,1,126}
\definecolor{n4}{RGB}{122,1,119}

\definecolor{e12}{RGB}{140,81,10}
\definecolor{e13}{RGB}{150,150,150}
\definecolor{e14}{RGB}{150,150,150}
\definecolor{e23}{RGB}{128,205,193}
\definecolor{e24}{RGB}{90,180,172}
\definecolor{e34}{RGB}{1,102,94}

\definecolor{e21}{RGB}{140,81,10}
\definecolor{e31}{RGB}{216,179,101}
\definecolor{e41}{RGB}{246,232,195}
\definecolor{e32}{RGB}{150,150,150}
\definecolor{e42}{RGB}{150,150,150}
\definecolor{e43}{RGB}{1,102,94}

\definecolor{c00}{RGB}{150,150,150}

\newcommand\va{0}
\newcommand\vb{1}
\newcommand\vc{2}
\newcommand\vd{3}

\newcommand{\nodes}[5]{
\node[circle,draw,fill=n1,inner sep=0pt,minimum size=1.2*\grid] (x1) at ($(#5) + (#1)$) {\scriptsize 1};
\node[circle,draw,fill=n2,inner sep=0pt,minimum size=1.2*\grid] (x2) at ($(#5) + (#2)$) {\scriptsize 2};
\node[circle,draw,fill=n3,inner sep=0pt,minimum size=1.2*\grid] (x3) at ($(#5) + (#3)$) {\scriptsize 3};
\node[circle,draw,fill=n4,inner sep=0pt,minimum size=1.2*\grid] (x4) at ($(#5) + (#4)$) {\scriptsize 4};
}

\newcommand{\edges}[8]{
\draw[->,-stealth,draw=e12] (x1) to[bend #1] node[left]{} (x2);
\draw[->,-stealth,draw=e23] (x2) to[bend #2] node[left]{} (x3);
\draw[->,-stealth,draw=e24] (x2) to[bend #3] node[left]{} (x4);
\draw[->,-stealth,draw=e34] (x3) to[bend #4] node[left]{} (x4);
\draw[->,-stealth,draw=e21] (x2) to[bend #5] node[left]{} (x1);
\draw[->,-stealth,draw=e31] (x3) to[bend #6] node[left]{} (x1);
\draw[->,-stealth,draw=e41] (x4) to[bend #7] node[left]{} (x1);
\draw[->,-stealth,draw=e43] (x4) to[bend #8] node[left]{} (x3);
}

\newcommand{\representation}[7]{
\coordinate (p1) at ($(#5) + (-3.0,+8.3)$);
\coordinate (p2) at ($(#5) + (-1.0,+8.3)$);

\draw[step=\grid, black] ($(p1) + (0,0)$) grid ($(p1) + (1,-4)$) node[label={[xshift=-0.4*\grid, yshift=2.6*\grid]\footnotesize #6}] {};
\draw[step=\grid, black] ($(p2) + (0,0)$) grid ($(p2) + (4,-4)$) node[label={[xshift=-1.6*\grid, yshift=2.6*\grid]\footnotesize #7}] {};

\draw[draw=black,fill=n1] ($(p1) + (0,-{#1})$) rectangle ($(p1) + (1,-{#1}) + (0,-1)$);
\draw[draw=black,fill=n2] ($(p1) + (0,-{#2})$) rectangle ($(p1) + (1,-{#2}) + (0,-1)$);
\draw[draw=black,fill=n3] ($(p1) + (0,-{#3})$) rectangle ($(p1) + (1,-{#3}) + (0,-1)$);
\draw[draw=black,fill=n4] ($(p1) + (0,-{#4})$) rectangle ($(p1) + (1,-{#4}) + (0,-1)$);

\draw[draw=black,fill=c00] ($(p2) + ({#1},-{#1})$) rectangle ($(p2) + ({#1},-{#1}) + (1,-1)$);
\draw[draw=black,fill=e21] ($(p2) + ({#1},-{#2})$) rectangle ($(p2) + ({#1},-{#2}) + (1,-1)$);
\draw[draw=black,fill=e31] ($(p2) + ({#1},-{#3})$) rectangle ($(p2) + ({#1},-{#3}) + (1,-1)$);
\draw[draw=black,fill=e41] ($(p2) + ({#1},-{#4})$) rectangle ($(p2) + ({#1},-{#4}) + (1,-1)$);

\draw[draw=black,fill=e12] ($(p2) + ({#2},-{#1})$) rectangle ($(p2) + ({#2},-{#1}) + (1,-1)$);
\draw[draw=black,fill=c00] ($(p2) + ({#2},-{#2})$) rectangle ($(p2) + ({#2},-{#2}) + (1,-1)$);
\draw[draw=black,fill=e32] ($(p2) + ({#2},-{#3})$) rectangle ($(p2) + ({#2},-{#3}) + (1,-1)$);
\draw[draw=black,fill=e42] ($(p2) + ({#2},-{#4})$) rectangle ($(p2) + ({#2},-{#4}) + (1,-1)$);

\draw[draw=black,fill=e13] ($(p2) + ({#3},-{#1})$) rectangle ($(p2) + ({#3},-{#1}) + (1,-1)$);
\draw[draw=black,fill=e23] ($(p2) + ({#3},-{#2})$) rectangle ($(p2) + ({#3},-{#2}) + (1,-1)$);
\draw[draw=black,fill=c00] ($(p2) + ({#3},-{#3})$) rectangle ($(p2) + ({#3},-{#3}) + (1,-1)$);
\draw[draw=black,fill=e43] ($(p2) + ({#3},-{#4})$) rectangle ($(p2) + ({#3},-{#4}) + (1,-1)$);

\draw[draw=black,fill=e14] ($(p2) + ({#4},-{#1})$) rectangle ($(p2) + ({#4},-{#1}) + (1,-1)$);
\draw[draw=black,fill=e24] ($(p2) + ({#4},-{#2})$) rectangle ($(p2) + ({#4},-{#2}) + (1,-1)$);
\draw[draw=black,fill=e34] ($(p2) + ({#4},-{#3})$) rectangle ($(p2) + ({#4},-{#3}) + (1,-1)$);
\draw[draw=black,fill=c00] ($(p2) + ({#4},-{#4})$) rectangle ($(p2) + ({#4},-{#4}) + (1,-1)$);
}

\newcommand{\subgraphtargetone}[6]{
\node[circle,draw,fill=c3,inner sep=0pt,minimum size=1.2*\grid] (x2) at ($(#1) + (#2)$) {\scriptsize2};

\draw[->,-stealth,draw=c3] (x1) to[bend #3] node[left]{} (x2);
\draw[->,-stealth,draw=c3] (x2) to[bend #4] node[left]{} (x3);
\draw[->,-stealth,draw=c3] (x2) to[bend #5] node[left]{} (x4);
\draw[->,-stealth,draw=c3] (x2) to[bend #6] node[left]{} (x1);
}

\newcommand{\representationtargetone}[5]{
\coordinate (p1) at ($(#5) + (-3.0,+8.3)$);
\coordinate (p2) at ($(#5) + (-1.0,+8.3)$);

\draw[draw=black,fill=c3] ($(p1) + (0,-{#2})$) rectangle ($(p1) + (1,-{#2}) + (0,-1)$);

\draw[draw=black,fill=c3] ($(p2) + ({#1},-{#2})$) rectangle ($(p2) + ({#1},-{#2}) + (1,-1)$);

\draw[draw=black,fill=c3] ($(p2) + ({#2},-{#1})$) rectangle ($(p2) + ({#2},-{#1}) + (1,-1)$);
\draw[draw=black,fill=c3] ($(p2) + ({#2},-{#2})$) rectangle ($(p2) + ({#2},-{#2}) + (1,-1)$);
\draw[draw=black,fill=c3] ($(p2) + ({#2},-{#3})$) rectangle ($(p2) + ({#2},-{#3}) + (1,-1)$);
\draw[draw=black,fill=c3] ($(p2) + ({#2},-{#4})$) rectangle ($(p2) + ({#2},-{#4}) + (1,-1)$);

\draw[draw=black,fill=c3] ($(p2) + ({#3},-{#2})$) rectangle ($(p2) + ({#3},-{#2}) + (1,-1)$);

\draw[draw=black,fill=c3] ($(p2) + ({#4},-{#2})$) rectangle ($(p2) + ({#4},-{#2}) + (1,-1)$);
}

\newcommand{\nodesmarginal}[5]{
\node[circle,draw,fill=n1,inner sep=0pt,minimum size=1.2*\grid] (x1) at ($(#5) + (#1)$) {\scriptsize 1};
% \node[circle,draw,fill=n2,inner sep=0pt,minimum size=1.2*\grid] (x2) at ($(#5) + (#2)$) {\scriptsize2};
\node[circle,draw,fill=n3,inner sep=0pt,minimum size=1.2*\grid] (x3) at ($(#5) + (#3)$) {\scriptsize 3};
\node[circle,draw,fill=n4,inner sep=0pt,minimum size=1.2*\grid] (x4) at ($(#5) + (#4)$) {\scriptsize 4};
}

\newcommand{\edgesmarginal}[8]{
% \draw[->,-stealth,draw=e12] (x1) to[bend #1] node[left]{} (x2);
% \draw[->,-stealth,draw=e23] (x2) to[bend #2] node[left]{} (x3);
% \draw[->,-stealth,draw=e24] (x2) to[bend #3] node[left]{} (x4);
\draw[->,-stealth,draw=e34] (x3) to[bend #4] node[left]{} (x4);
% \draw[->,-stealth,draw=e21] (x2) to[bend #5] node[left]{} (x1);
\draw[->,-stealth,draw=e31] (x3) to[bend #6] node[left]{} (x1);
\draw[->,-stealth,draw=e41] (x4) to[bend #7] node[left]{} (x1);
\draw[->,-stealth,draw=e43] (x4) to[bend #8] node[left]{} (x3);
}

\newcommand{\representationmarginal}[7]{
\coordinate (p1) at ($(#5) + (-3.0,+8.3)$);
\coordinate (p2) at ($(#5) + (-1.0,+8.3)$);

\draw[step=\grid, black] ($(p1) + (0,0)$) grid ($(p1) + (1,-3)$) node[label={[xshift=-0.45*\grid, yshift=1.8*\grid]\footnotesize #6}] {};
\draw[step=\grid, black] ($(p2) + (0,0)$) grid ($(p2) + (3,-3)$) node[label={[xshift=-1.25*\grid, yshift=1.8*\grid]\footnotesize #7}] {};

\draw[draw=black,fill=n1] ($(p1) + (0,-{#1})$) rectangle ($(p1) + (1,-{#1}) + (0,-1)$);
\draw[draw=black,fill=n3] ($(p1) + (0,-{#2})$) rectangle ($(p1) + (1,-{#2}) + (0,-1)$);
\draw[draw=black,fill=n4] ($(p1) + (0,-{#3})$) rectangle ($(p1) + (1,-{#3}) + (0,-1)$);

\draw[draw=black,fill=c00] ($(p2) + ({#1},-{#1})$) rectangle ($(p2) + ({#1},-{#1}) + (1,-1)$);
\draw[draw=black,fill=e31] ($(p2) + ({#1},-{#2})$) rectangle ($(p2) + ({#1},-{#2}) + (1,-1)$);
\draw[draw=black,fill=e41] ($(p2) + ({#1},-{#3})$) rectangle ($(p2) + ({#1},-{#3}) + (1,-1)$);

\draw[draw=black,fill=e13] ($(p2) + ({#2},-{#1})$) rectangle ($(p2) + ({#2},-{#1}) + (1,-1)$);
\draw[draw=black,fill=c00] ($(p2) + ({#2},-{#2})$) rectangle ($(p2) + ({#2},-{#2}) + (1,-1)$);
\draw[draw=black,fill=e43] ($(p2) + ({#2},-{#3})$) rectangle ($(p2) + ({#2},-{#3}) + (1,-1)$);

\draw[draw=black,fill=e14] ($(p2) + ({#3},-{#1})$) rectangle ($(p2) + ({#3},-{#1}) + (1,-1)$);
\draw[draw=black,fill=e34] ($(p2) + ({#3},-{#2})$) rectangle ($(p2) + ({#3},-{#2}) + (1,-1)$);
\draw[draw=black,fill=c00] ($(p2) + ({#3},-{#3})$) rectangle ($(p2) + ({#3},-{#3}) + (1,-1)$);
}

\hspace{-90pt}
\begin{tikzpicture}[font=\footnotesize,x=\grid,y=\grid,scale=0.8]
\coordinate (ua) at (-1.9,11.0);
\coordinate (ub) at (+2.0,12.0);
\coordinate (uc) at (+0.0,13.5);
\coordinate (ud) at (-0.5,15.5);

\coordinate (a)  at (-35,9);
\coordinate (b)  at (-12,0);
\coordinate (c)  at ( 14,0);

\node at (a) {\scriptsize
$\begin{aligned}
    h(\mathbf{G}^3_a,\textcolor{c3}{\mathbf{G}^1_b};4)
    &=
    \prod_{t\notin\mathbf{s}}\mathds{1}_{\mathbf{x}_t}(\mathbf{X}_t)\prod_{u\notin\mathbf{s}}\mathds{1}_{\mathbf{a}_{tu}}(\mathbf{A}_{tu})\mathds{1}_{\mathsf{dom}(\textcolor{c3}{\mathbf{X}_2)}}(\textcolor{c3}{\mathbf{X}_2})
    \\
    &\hspace{-40pt}\times
    \mathds{1}_{\mathsf{dom}(\textcolor{c3}{\mathbf{A}_{22}})}(\textcolor{c3}{\mathbf{A}_{22}})\prod_{k\notin\mathbf{s}}\mathds{1}_{\mathsf{dom}(\textcolor{c3}{\mathbf{A}_{2k})}}(\textcolor{c3}{\mathbf{A}_{2k}})\mathds{1}_{\mathsf{dom}(\textcolor{c3}{\mathbf{A}_{k2})}}(\textcolor{c3}{\mathbf{A}_{k2}})
\end{aligned}$
};
\node at ($(a) + (2.0,-6.5)$) {(a) An example of \eqref{eq:h} for $\mathbf{s}\coloneqq\{2\}$};

\nodes{ua}{ub}{uc}{ud}{b}
\edges{left}{right}{right}{right}{right}{right}{right}{left}
\representation{\va}{\vb}{\vc}{\vd}{b}{$\mathbf{X}$}{$\mathbf{A}$}
\node at ($(b) + (1.0,2.5)$) {(b) The corresponding node of $\mathbf{G}$};

\representationtargetone{\va}{\vb}{\vc}{\vd}{b}
\subgraphtargetone{b}{ub}{left}{right}{right}{right}
\nodesmarginal{ua}{ub}{uc}{ud}{c}
\edgesmarginal{left}{right}{right}{right}{right}{right}{right}{left}
\representationmarginal{\va}{\vb}{\vc}{\vd}{c}{$\mathbf{x}$}{$\mathbf{a}$}
\node at ($(c) + (-2.0,2.5)$) {(c) $\mathbf{G}$ after the marginal evidence query};

\end{tikzpicture}}
        \caption{\emph{An example of the marginal evidence query over a 4-node graph.} (a) An instantiation of the omni-compatible GC \eqref{eq:h} over a graph $\mathbf{G}^4$ with targetting the 2nd node, $\mathbf{s}\coloneqq\{2\}$. The orange color highlights the targeted node and its associated edges. (b) A visual representation of $\mathbf{G}$, where the targeted node and its associated edges, which correspond to (a), are highlighted in orange. (c) After performing the marginal evidence query over the 2nd node of $\mathbf{G}^4$, we obtain a new 3-node evidence graph $\mathbf{g}$.}   
    \label{fig:inference}
    \end{center}
\end{figure*}

\textbf{Tractable expectation.} Most inference queries can collectively be expressed in terms of the following expectation:
\begin{align}\label{eq:expectation}
    \mathbb{E}_{p(\mathbf{G})}[h(\mathbf{G})]
    % &\coloneqq
    % \int h(\mathbf{g})p(\mathbf{g})d\mathbf{g}
    % \nonumber
    % \\
    &=
    \sum^\infty_{n=1}\int h(\mathbf{g}^n)p(\mathbf{g}^n|n)p(n)d\mathbf{g}^n
    ,
\end{align}
where $h$ is a GC (Definition \ref{def:graph-circuit}) whose specific form allows us to formulate a desired query over (a part of) $\mathbf{G}$.

Before focusing on specific instances of \eqref{eq:expectation}, we introduce the conditions for its tractability in \cref{prop:tractability-expectation}.

\begin{proposition}{(Tractable expectation of a GC.)}\label{prop:tractability-expectation}
    Let $h$ and $p$ be two tractable (\cref{prop:tractability-pgc}) and compatible (\cref{ass:compatibility}) (P)GCs, such that the product of any two input units $u\in h$ and $v\in p$ with the same scope $\mathbf{G}_u=\mathbf{G}_v$ (\cref{def:graph-scope}) has a tractable integral; then, the expectation \eqref{eq:expectation} is tractable.
\end{proposition}

Apart from \cref{prop:tractability-pgc}, the key requirement for the tractability of \eqref{eq:expectation} is the\emph{compatibility} of $h$ and $p$. The standard definition of compatibility (\cref{ass:compatibility}) does not change under our variable-size (\ref{challenges:random-size}) and $\mathbb{S}_n$-invariant (\ref{challenges:symmetry}) setting since $\mathbf{g}^n$ enters both $h$ and $p$ as the same realization (with a fixed size and values). Note that the part of \cref{prop:tractability-pgc} requiring the restricted support of the cardinality distribution can be relaxed for $h$, as it is satisfied by $p$.

\textbf{Querying $k$-node subgraphs.} There can be many examples of a GC, $h$, that is compatible with a PGC, $p$. We will focus on a simple, yet very powerful, omni-compatible \cite{vergari2021compositional}, and tractably $\mathbb{S}_n$-invariant circuit, which will allow us to formulate many desired inference queries.

\begin{proposition}{(Omni-Compatible Tractably $\mathbb{S}_n$-invariant GC.)}\label{prop:omni-compatible-graph-circuit}
    Let $\mathbf{G}^n=\lbrace\mathbf{G}^{n-k}_a,\mathbf{G}^k_b\rbrace$, where $\mathbf{G}^{n-k}_a$ and $\mathbf{G}^k_b$ are connected $(n\text{-}k)$-node and $k$-node subgraphs of $\mathbf{G}^n$. A fully factorized, $n$-dependent, GC of the following form:
    \begin{align}\label{eq:h}
        \displaystyle
        h(\mathbf{G}^{n-k}_a,\mathbf{G}^k_b;n)
        &\coloneqq
        \prod_{t\notin\mathbf{s}}h(\mathbf{X}_t)
        \hspace{-2pt}
        \prod_{u\notin\mathbf{s}}h(\mathbf{A}_{tu})
        \nonumber
        \\
        &\hspace{-43pt}\times
        \prod_{i\in\mathbf{s}}h(\mathbf{X}_i)
        \prod_{j\in\mathbf{s}}h(\mathbf{A}_{ij})
        \prod_{l\notin\mathbf{s}}h(\mathbf{A}_{il})h(\mathbf{A}_{li})
    \end{align}
    is tractably $\mathbb{S}_n$-invariant and omni-compatible. $\mathbf{s}\coloneqq\{s_1,\ldots,s_k\}\subseteq[n]$ are indices of $k$ nodes of $\mathbf{G}^n$ belonging to $\mathbf{G}^k_b$.
\end{proposition}

The omni-compatible GC \eqref{eq:h} is simply the product of functions over all node and edge features in $\mathbf{G}^n$, which is rearranged into two sets $\mathbf{G}^{n-k}_a$ and~$\mathbf{G}^k_b$ (\cref{sec:proof-omni-compatible-graph-circuit}).

\textbf{Targeting nodes and (or) edges.} The utility of \cref{prop:omni-compatible-graph-circuit} lies in that a suitable choice of $h(\mathbf{X}_i)$ and $h(\mathbf{A}_{ij})$ allows us to target $i$-th node of $\mathbf{G}$ and the edge between $i$-th and $j$-th node of $\mathbf{G}$, respectively. These functions facilitate the formulation of various queries of interest. For example, if $h(\mathbf{X}_i)\coloneqq\mathds{1}_\mathcal{X}(\mathbf{X}_i)$, where $\mathds{1}_\mathcal{X}$ is the indicator function, and $h(\mathbf{A}_{ij})\coloneqq\mathds{1}_\mathcal{A}(\mathbf{A}_{ij})$, we can obtain basic inference queries just by a mere choice of the sets $\mathcal{X}$ and $\mathcal{A}$. The \emph{evidence} query is computed for $\mathcal{X}\coloneqq\mathbf{x}_i\in\mathsf{dom}(\mathbf{X}_i)$ and $\mathcal{A}\coloneqq\mathbf{a}_{ij}\in\mathsf{dom}(\mathbf{A}_{ij})$. The \emph{marginal} query is given by $\mathcal{X}\coloneqq\mathsf{dom}(\mathbf{X}_i)$ and $\mathcal{A}\coloneqq\mathsf{dom}(\mathbf{A}_{ij})$. $r$-th \emph{moment} query results from $h(\mathbf{X}_i)\coloneqq\mathbf{X}^{r_i}_i$ and $h(\mathbf{A}_{ij})\coloneqq\mathbf{A}^{r_{ij}}_{ij}$, where $r_i$ and $r_{ij}$ are non-negative integers. We demonstrate an example of the marginal query for $\mathbf{s}\coloneqq\{2\}$ in \cref{fig:inference}.

\section{Proofs}
\label{sec:proofs}

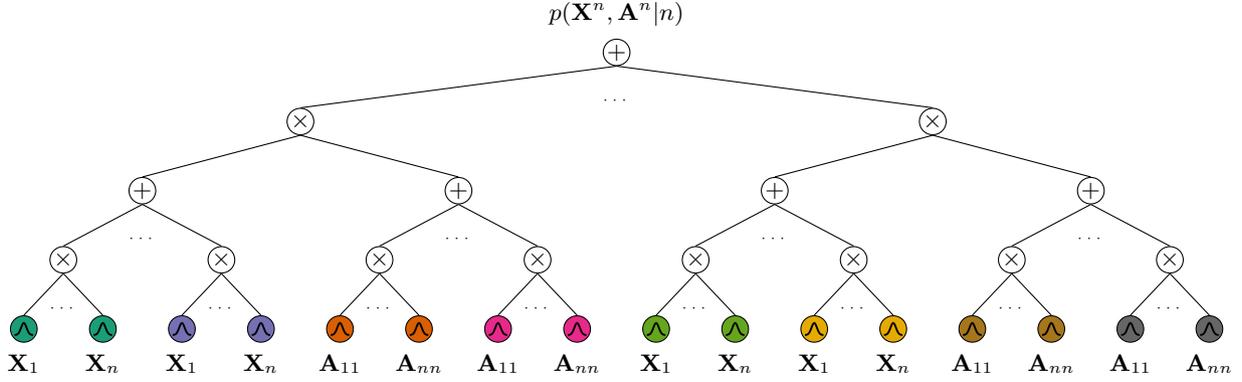
\begin{figure*}[!t]
    \begin{center}
        \centerline{\tikzset{
    snode/.style={
        font={\small $+$},
        circle,
        draw,
        align=center,
        inner sep=0pt,
        minimum size=10pt,
    },
    pnode/.style={
        font={\small $\times$},
        circle,
        draw,
        align=center,
        inner sep=0pt,
        minimum size=10pt,
    },
    lnode/.style={
        font={%
            \begin{tikzpicture}[font=\small, >=stealth,yscale=0.40,xscale=0.18]
                \draw[line width=0.25mm,domain=-1.7:1.7] plot (\x, {exp((-(\x)^2)/0.5)});
            \end{tikzpicture}%
            },
        circle,
        draw,
        align=center,
        inner sep=0pt,
        minimum size=10pt,
    },
}

\tikzset{
    inode/.style={
        inner sep=0pt,
        text width=1pt,
    },
    hnode/.style={
        font={\scriptsize $\hdots$},
        draw=none, 
        align=center,
        inner sep=0pt,
        minimum size=10pt,
    },
    enode/.style={
        font={},
        circle,
        draw,
        dotted,
        align=center,
        inner sep=0pt,
        minimum size=10pt,
  },
}

\begin{tikzpicture}[font=\small, x=0.4cm, y=0.4cm, >={Triangle[scale=0.7pt]}]
    \definecolor{c0}{RGB}{243,246,249}
    \definecolor{c1}{RGB}{27,158,119}
    \definecolor{c2}{RGB}{117,112,179}
    \definecolor{c3}{RGB}{217,95,2}
    \definecolor{c4}{RGB}{231,41,138}
    \definecolor{c5}{RGB}{102,166,30}
    \definecolor{c6}{RGB}{230,171,2}
    \definecolor{c7}{RGB}{166,118,29}
    \definecolor{c8}{RGB}{102,102,102}

    \coordinate (a) at (-2.2,0);
    \coordinate (b) at ( 2.2,0);

    \def\cx{+1.30}
    \def\cy{+2.30}

    \node [snode, label={[xshift=0.0pt, yshift=1.4pt] $p(\mathbf{X}^n,\mathbf{A}^n|n)$}] (root) at (a) {};

    \node [pnode] (p1_2) at ($(a) + (-8*\cx,-1*\cy)$) {};
    \node [pnode] (p2_2) at ($(a) + ( 8*\cx,-1*\cy)$) {};

    \node [snode] (s1_3) at ($(a) + (-12*\cx,-2*\cy)$) {};
    \node [snode] (s2_3) at ($(a) + ( -4*\cx,-2*\cy)$) {};
    \node [snode] (s3_3) at ($(a) + (  4*\cx,-2*\cy)$) {};
    \node [snode] (s4_3) at ($(a) + ( 12*\cx,-2*\cy)$) {};

    \node [pnode] (p1_4) at ($(a) + (-14*\cx,-3*\cy)$) {};
    \node [pnode] (p2_4) at ($(a) + (-10*\cx,-3*\cy)$) {};
    \node [pnode] (p3_4) at ($(a) + ( -6*\cx,-3*\cy)$) {};
    \node [pnode] (p4_4) at ($(a) + ( -2*\cx,-3*\cy)$) {};
    \node [pnode] (p5_4) at ($(a) + (  2*\cx,-3*\cy)$) {};
    \node [pnode] (p6_4) at ($(a) + (  6*\cx,-3*\cy)$) {};
    \node [pnode] (p7_4) at ($(a) + ( 10*\cx,-3*\cy)$) {};
    \node [pnode] (p8_4) at ($(a) + ( 14*\cx,-3*\cy)$) {};

    \node [lnode,fill=c1] (l1_5)  at ($(a) + (-15*\cx,-4*\cy)$) {};
    \node [lnode,fill=c1] (l2_5)  at ($(a) + (-13*\cx,-4*\cy)$) {};
    \node [lnode,fill=c2] (l3_5)  at ($(a) + (-11*\cx,-4*\cy)$) {};
    \node [lnode,fill=c2] (l4_5)  at ($(a) + ( -9*\cx,-4*\cy)$) {};
    \node [lnode,fill=c3] (l5_5)  at ($(a) + ( -7*\cx,-4*\cy)$) {};
    \node [lnode,fill=c3] (l6_5)  at ($(a) + ( -5*\cx,-4*\cy)$) {};
    \node [lnode,fill=c4] (l7_5)  at ($(a) + ( -3*\cx,-4*\cy)$) {};
    \node [lnode,fill=c4] (l8_5)  at ($(a) + ( -1*\cx,-4*\cy)$) {};
    \node [lnode,fill=c5] (l9_5)  at ($(a) + (  1*\cx,-4*\cy)$) {};
    \node [lnode,fill=c5] (l10_5) at ($(a) + (  3*\cx,-4*\cy)$) {};
    \node [lnode,fill=c6] (l11_5) at ($(a) + (  5*\cx,-4*\cy)$) {};
    \node [lnode,fill=c6] (l12_5) at ($(a) + (  7*\cx,-4*\cy)$) {};
    \node [lnode,fill=c7] (l13_5) at ($(a) + (  9*\cx,-4*\cy)$) {};
    \node [lnode,fill=c7] (l14_5) at ($(a) + ( 11*\cx,-4*\cy)$) {};
    \node [lnode,fill=c8] (l15_5) at ($(a) + ( 13*\cx,-4*\cy)$) {};
    \node [lnode,fill=c8] (l16_5) at ($(a) + ( 15*\cx,-4*\cy)$) {};

    \node [inode, label={\small $\mathbf{X}_1$}   ] (x1) at ($(a) + (-15*\cx,-4.8*\cy)$) {};
    \node [inode, label={\small $\mathbf{X}_n$}   ] (x1) at ($(a) + (-13*\cx,-4.8*\cy)$) {};
    \node [inode, label={\small $\mathbf{X}_1$}   ] (x1) at ($(a) + (-11*\cx,-4.8*\cy)$) {};
    \node [inode, label={\small $\mathbf{X}_n$}   ] (x1) at ($(a) + ( -9*\cx,-4.8*\cy)$) {};
    \node [inode, label={\small $\mathbf{A}_{11}$}] (x1) at ($(a) + ( -7*\cx,-4.8*\cy)$) {};
    \node [inode, label={\small $\mathbf{A}_{nn}$}] (x1) at ($(a) + ( -5*\cx,-4.8*\cy)$) {};
    \node [inode, label={\small $\mathbf{A}_{11}$}] (x1) at ($(a) + ( -3*\cx,-4.8*\cy)$) {};
    \node [inode, label={\small $\mathbf{A}_{nn}$}] (x1) at ($(a) + ( -1*\cx,-4.8*\cy)$) {};
    \node [inode, label={\small $\mathbf{X}_1$}   ] (x1) at ($(a) + (  1*\cx,-4.8*\cy)$) {};
    \node [inode, label={\small $\mathbf{X}_n$}   ] (x1) at ($(a) + (  3*\cx,-4.8*\cy)$) {};
    \node [inode, label={\small $\mathbf{X}_1$}   ] (x1) at ($(a) + (  5*\cx,-4.8*\cy)$) {};
    \node [inode, label={\small $\mathbf{X}_n$}   ] (x1) at ($(a) + (  7*\cx,-4.8*\cy)$) {};
    \node [inode, label={\small $\mathbf{A}_{11}$}] (x1) at ($(a) + (  9*\cx,-4.8*\cy)$) {};
    \node [inode, label={\small $\mathbf{A}_{nn}$}] (x1) at ($(a) + ( 11*\cx,-4.8*\cy)$) {};
    \node [inode, label={\small $\mathbf{A}_{11}$}] (x1) at ($(a) + ( 13*\cx,-4.8*\cy)$) {};
    \node [inode, label={\small $\mathbf{A}_{nn}$}] (x1) at ($(a) + ( 15*\cx,-4.8*\cy)$) {};

    \draw [-] (root.south) -- (p1_2.north) node[midway,above left]  {};
    \draw [-] (root.south) -- (p2_2.north) node[midway,above right] {};

    \draw [-] (p1_2.south) -- (s1_3.north) node[midway,above left]  {};
    \draw [-] (p1_2.south) -- (s2_3.north) node[midway,above right] {};
    \draw [-] (p2_2.south) -- (s3_3.north) node[midway,above left]  {};
    \draw [-] (p2_2.south) -- (s4_3.north) node[midway,above right] {};

    \draw [-] (s1_3.south) -- (p1_4.north) node[midway,above left]  {};
    \draw [-] (s1_3.south) -- (p2_4.north) node[midway,above right] {};
    \draw [-] (s2_3.south) -- (p3_4.north) node[midway,above left]  {};
    \draw [-] (s2_3.south) -- (p4_4.north) node[midway,above right] {};
    \draw [-] (s3_3.south) -- (p5_4.north) node[midway,above left]  {};
    \draw [-] (s3_3.south) -- (p6_4.north) node[midway,above right] {};
    \draw [-] (s4_3.south) -- (p7_4.north) node[midway,above left]  {};
    \draw [-] (s4_3.south) -- (p8_4.north) node[midway,above right] {};

    \draw [-] (p1_4.south) -- (l1_5.north);
    \draw [-] (p1_4.south) -- (l2_5.north);
    \draw [-] (p2_4.south) -- (l3_5.north);
    \draw [-] (p2_4.south) -- (l4_5.north);
    \draw [-] (p3_4.south) -- (l5_5.north);
    \draw [-] (p3_4.south) -- (l6_5.north);
    \draw [-] (p4_4.south) -- (l7_5.north);
    \draw [-] (p4_4.south) -- (l8_5.north);
    \draw [-] (p5_4.south) -- (l9_5.north);
    \draw [-] (p5_4.south) -- (l10_5.north);
    \draw [-] (p6_4.south) -- (l11_5.north);
    \draw [-] (p6_4.south) -- (l12_5.north);
    \draw [-] (p7_4.south) -- (l13_5.north);
    \draw [-] (p7_4.south) -- (l14_5.north);
    \draw [-] (p8_4.south) -- (l15_5.north);
    \draw [-] (p8_4.south) -- (l16_5.north);

    \node [inode, label={\tiny $\cdots$}] (d1) at ($(a) + (  0*\cx,-0.9*\cy)$) {};

    \node [inode, label={\tiny $\cdots$}] (d1) at ($(a) + (-12*\cx,-2.9*\cy)$) {};
    \node [inode, label={\tiny $\cdots$}] (d1) at ($(a) + ( -4*\cx,-2.9*\cy)$) {};
    \node [inode, label={\tiny $\cdots$}] (d1) at ($(a) + (  4*\cx,-2.9*\cy)$) {};
    \node [inode, label={\tiny $\cdots$}] (d1) at ($(a) + ( 12*\cx,-2.9*\cy)$) {};

    \node [inode, label={\tiny $\cdots$}] (d1) at ($(a) + (-14*\cx,-3.9*\cy)$) {};
    \node [inode, label={\tiny $\cdots$}] (d1) at ($(a) + (-10*\cx,-3.9*\cy)$) {};
    \node [inode, label={\tiny $\cdots$}] (d1) at ($(a) + ( -6*\cx,-3.9*\cy)$) {};
    \node [inode, label={\tiny $\cdots$}] (d1) at ($(a) + ( -2*\cx,-3.9*\cy)$) {};
    \node [inode, label={\tiny $\cdots$}] (d1) at ($(a) + (  2*\cx,-3.9*\cy)$) {};
    \node [inode, label={\tiny $\cdots$}] (d1) at ($(a) + (  6*\cx,-3.9*\cy)$) {};
    \node [inode, label={\tiny $\cdots$}] (d1) at ($(a) + ( 10*\cx,-3.9*\cy)$) {};
    \node [inode, label={\tiny $\cdots$}] (d1) at ($(a) + ( 14*\cx,-3.9*\cy)$) {};

\end{tikzpicture}}
        \caption{\emph{An inherently $\mathbb{S}_n$-invariant PGC through the conditional i.i.d.\ assumption.} The $n$-conditioned part of a PGC that is tractable and inherently $\mathbb{S}_n$-invariant, as formulated in \cref{prop:factorized-pgcs}. The input units with the same color share the parameterization and correspond to the product terms in \eqref{eq:iid-components}.}
        \label{fig:inherently-sn-invariant-pgc}
    \end{center}
\end{figure*}

\subsection{Proof of Proposition \ref{prop:factorized-pgcs}}
\label{sec:proof-factorized-pgcs}
To prove that the PGC from \cref{prop:factorized-pgcs} is inherently $\mathbb{S}_n$-invariant, we need to show that the components \eqref{eq:iid-components} are $\mathbb{S}_n$-invariant according to the definition of the tractable $\mathbb{S}_n$-invariance in \cref{def:tractability-sn-invariance}. The i.i.d.\ assumption means that the parameters of the product terms in \ref{eq:iid-components} are identical. To make the explicit, let us write
\begin{equation}\nonumber
    p(\mathbf{G}^n|\mathbf{z},n)
    =
    \prod_{i\in[n]}p(\mathbf{X}_i|\mathbf{z},n;\theta_X)\prod_{j\in[n]}p(\mathbf{A}_{ij}|\mathbf{z},n;\theta_A)
    ,
\end{equation}
where $\theta_X$ and $\theta_A$ are parameters associated with the nodes and edges, respectively. It is easy to see now that
\begin{equation}\nonumber
    \prod_{i\in[n]}p(\mathbf{X}_{\pi(i)}|\mathbf{z},n;\theta_X)\prod_{j\in[n]}p(\mathbf{A}_{\pi(i)\pi(j)}|\mathbf{z},n;\theta_A)
    =
    \prod_{i\in[n]}p(\mathbf{X}_i|\mathbf{z},n;\theta_X)\prod_{j\in[n]}p(\mathbf{A}_{ij}|\mathbf{z},n;\theta_A)
    ,
\end{equation}
for all $\bm{\pi}\in\mathbb{S}_n$. Since this invariance also holds for all $\mathbf{z}\in\mathsf{dom}(\mathbf{Z})$ and propagates through the summation in \eqref{eq:latent-representation}, we can conclude that the PGC from \cref{prop:factorized-pgcs} is inherently $\mathbb{S}_n$-invariant. \qed

\subsection{Proof of Proposition \ref{prop:omni-compatible-graph-circuit}}
\label{sec:proof-omni-compatible-graph-circuit}
To demonstrate that $h(\mathbf{G}^n;n)$ is inherently $\mathbb{S}_n$-invariant, let us first make the parameterization of the individual terms $h(\mathbf{X}_i)$ and $h(\mathbf{A}_{ij})$ explicit as follows:
\begin{equation}\nonumber
    h(\mathbf{G}^n;n)
    \coloneqq
    \prod_{i\in[n]}h(\mathbf{X}_i;\theta_i)\prod_{j\in[n]}h(\mathbf{A}_{ij};\theta_{ij}),
\end{equation}
where $\theta_i$ and $\theta_{ij}$ are parameters associated to $\mathbf{X}_i$ and $\mathbf{A}_{ij}$, respectively. It holds that $h(\bm{\pi}\mathbf{G}^n;n)=h(\mathbf{G}^n;n)$, for all $\bm{\pi}\in\mathbb{S}_n$, only if $\theta_i\coloneqq\theta_X$ and $\theta_{ij}\coloneqq\theta_A$ for all $i\in[n]$ and $j\in[n]$,
\begin{equation}\nonumber
    \prod_{i\in[n]}h(\mathbf{X}_{\pi(i)};\theta_X)\prod_{j\in[n]}h(\mathbf{A}_{\pi(i)\pi(j)};\theta_A)
    =
    \prod_{i\in[n]}h(\mathbf{X}_i;\theta_X)\prod_{j\in[n]}h(\mathbf{A}_{ij};\theta_A).
\end{equation}
This fully factorized form is what allows us to write $h(\mathbf{G}^n;n)=h(\mathbf{G}_a^{n-k},\mathbf{G}_b^k;n)\coloneqq h(\mathbf{G}_a^{n-k};n-k)h(\mathbf{G}_b^k;k)$, where
\begin{align}
    h(\mathbf{G}_a^{n-k};n-k)
    &=
    \prod_{t\notin\mathbf{s}}h(\mathbf{X}_t;\theta_X)\prod_{u\notin\mathbf{s}}h(\mathbf{A}_{tu};\theta_A)
    ,
    \nonumber
    \\
    h(\mathbf{G}_b^k;k)
    &=
    \prod_{i\in\mathbf{s}}h(\mathbf{X}_i;\theta_X)\prod_{j\in\mathbf{s}}h(\mathbf{A}_{ij};\theta_A)\prod_{l\notin\mathbf{s}}h(\mathbf{A}_{il};\theta_A)h(\mathbf{A}_{li};\theta_A)
    .
    \nonumber
\end{align}
Here, we can see that $\mathbf{G}^{n-k}_a\coloneqq\lbrace\mathbf{X}_t,\mathbf{A}_{tu}\rbrace^{}_{u,t\notin\mathbf{s}}$ and $\mathbf{G}^k_b\coloneqq\lbrace\mathbf{X}_i,\mathbf{A}_{ij}\rbrace^{}_{i,j\in\mathbf{s}}\cup\lbrace\mathbf{A}_{il},\mathbf{A}_{li}\rbrace_{i,l\notin\mathbf{s}}$.

The tractability of this $\mathbb{S}_n$-invariant function follows from the fact that it does not yield an approximate value for any of the $n!$ configurations of $\mathbf{G}^n$ (satisfying \cref{def:tractability-pc}(A)), and from its linear complexity (satisfying \cref{def:tractability-pc}(B)).

A fully decomposable GC, $h(\mathbf{G}^n;n)$, is omni-compatible if it is compatible (\cref{ass:compatibility}) with any tractably $\mathbb{S}_n$-invariant (\cref{def:tractability-sn-invariance}) GC over $\mathbf{G}^n$. \qed

\section{Experimental Details}
\label{sec:experimental-details}

The results for the baselines in Table \ref{tab:generation-unco}(top) are obtained from \citep{jo2022score,kong2023autoregressive}.

\subsection{Datasets}
\label{sec:datasets}
QM9 contains around 134k stable small organic molecules of at most 9 atoms that can take 4 different types. ZINC250k contains around 250k drug-like molecules of at most 38 atoms that can take 9 different types. We use the RDKit library \citep{landrum2006rdkit} to first kekulize the molecules and then remove the hydrogen atoms. The final molecules contain only the single, double, and triple bonds. Since our objective is to test different canonical graph orderings, we randomly permute the atoms in each molecule before applying the canonicalization (sorting) in order let each ordering result from the same initial graph configuration. We randomly partition the datasets into $80$\%, $10$\%, and $10$\% for train, validation, and testing, respectively. We repeat all experiments for 5 different seeds, i.e., 5 different train, validation, and testing splits.

\begin{table*}[ht]
    \renewcommand{\arraystretch}{0.8}
    \caption{\emph{Statistics of the molecular datasets}. The meaning of the symbols is as follows: $I$ is the number of instances, $N$ is the number of atoms in a single molecule, $n_X$ is the number of atom types, $n_A$ is the number of bond types, including the empty bond.}
    \label{tab:dataset-statistics}
    \begin{center}
        \begin{small}
            \begin{tabular}{lllccccc}
\toprule
Dataset & $I$ & min $N$ & max $N$ & $n_X$ & $n_A$ \\
\midrule
QM9      & 133,885 & 1 & 9 & 4 & 3+1\\
Zinc250k & 249,455 & 6 & 38 & 9 & 3+1\\
\bottomrule
\end{tabular}

        \end{small}
    \end{center}
\end{table*}

\subsection{Training Setup}
% \label{sec:training-setup}
We train all $\bm{\pi}$PGCs variants by minimizing the negative log-likelihood for $40$ epochs. We use the ADAM optimizer \citep{kingma2014adam} with $256$ samples in the minibatch, step-size $\alpha=0.05$, and decay rates $\beta_1=0.9$ and $\beta_2=0.82$. All experiments are repeated $5$ times with different initialization of the model's parameters. Following the baselines, we sample 10000 molecules to compute the metrics introduced in Section \ref{sec:experiments}.

% \subsection{$\bm{\pi}$PGCs Variants}
\subsection{PGC Variants}
\label{sec:pgc-variants}
The variants of PGCs described in \cref{sec:experiments} are instantiated depending on a concrete form of the region graph that is used to build the node and edge PCs (\cref{fig:pgc}). The expressive power of these monolithic and tensorized variants of PCs is driven by different sets of hyper-parameters, which we summarize in \cref{tab:hyper-parameters}. All these PGCs use the input units as categorical distributions, and their cardinality distribution is also the categorical distribution (whose parameters are trained via gradient descent along with all the parameters of the node and edge PCs). We run each combination of the hyper-parameters in \cref{tab:hyper-parameters} for each canonical ordering in the following set: $\lbrace$Random, BFT, DFT, RCM, MCA $\rbrace$, as introduced in \cref{sec:orderings}. Following \citep{jo2022score}, we select the best models based on the lowest validation FCD.

\begin{table*}[ht]
    \renewcommand{\arraystretch}{0.8}
    \caption{\emph{Hyper-parameters of the $\bm{\pi}$PGCs variants}. The meaning of the hyper-parameters is as follows: $n_l$ is the number of layers, $n_S$ is the number of sum units, $n_I$ is the number of input units, $n_R$ is the number of repetitions \citep{peharz2020random}, and $n_c$ is the number of weights in the root sum unit, which controls the ability to capture the correlations between the nodes and edges (\cref{fig:pgc}). N and E stands for the node-PC and edge-PC, respectively.}
    \label{tab:hyper-parameters}
    \begin{center}
        \begin{small}
            \begin{tabular}{lllccccc}
\toprule
Dataset & $\bm{\pi}$PGC & & $n_l$ & $n_S$ & $n_I$ & $n_R$ & $n_c$ \\
\midrule
\multirow{10}{*}{QM9}      & \multirow{2}{*}{BT}   & N & $\{1, 2, 3\}$ & $\{16, 32, 64\}$ & $\{16, 32\}$ & -                & \multirow{2}{*}{$\{1, 4, 16, 64, 256, 512\}$} \\
                           &                       & E & $\{3, 4, 5\}$ & $\{16, 32, 64\}$ & $\{16, 32\}$ & -                & \\
                           & \multirow{2}{*}{LT}   & N & $\{1, 2, 3\}$ & $\{16, 32, 64\}$ & $\{16, 32\}$ & -                & \multirow{2}{*}{$\{1, 4, 16, 64, 256, 512\}$} \\
                           &                       & E & $\{3, 4, 5\}$ & $\{16, 32, 64\}$ & $\{16, 32\}$ & -                & \\
                           & \multirow{2}{*}{RT}   & N & $\{1, 2, 3\}$ & $\{16, 32, 64\}$ & $\{16, 32\}$ & $\{16, 32, 64\}$ & \multirow{2}{*}{$\{1, 4, 16, 64, 256, 512\}$} \\
                           &                       & E & $\{3, 4, 5\}$ & $\{16, 32, 64\}$ & $\{16, 32\}$ & $\{16, 32, 64\}$ & \\
                           & \multirow{2}{*}{RT-S} & N & $\{1, 2, 3\}$ & $\{16, 32, 64\}$ & $\{16, 32\}$ & $\{16, 32, 64\}$ & \multirow{2}{*}{$\{1, 4, 16, 64, 256, 512\}$} \\
                           &                       & E & $\{3, 4, 5\}$ & $\{16, 32, 64\}$ & $\{16, 32\}$ & $\{16, 32, 64\}$ & \\
                           & \multirow{2}{*}{HCLT} & N & -             & $\{64, 128, 256, 512\}$ & -            & -         & \multirow{2}{*}{$\{1, 4, 16, 64, 256, 512\}$} \\
                           &                       & E & -             & $\{64, 128, 256, 512\}$ & -            & -         & \\
\midrule
\multirow{10}{*}{Zinc250k} & \multirow{2}{*}{BT}   & N & $\{2, 3, 4\}$ & $\{16, 32, 64\}$ & $\{16, 32\}$ & -                & \multirow{2}{*}{$\{1, 4, 16, 64, 256, 512\}$} \\
                           &                       & E & $\{2, 4, 6\}$ & $\{16, 32, 64\}$ & $\{16, 32\}$ & -                & \\
                           & \multirow{2}{*}{LT}   & N & $\{2, 3, 4\}$ & $\{16, 32, 64\}$ & $\{16, 32\}$ & -                & \multirow{2}{*}{$\{1, 4, 16, 64, 256, 512\}$} \\
                           &                       & E & $\{2, 4, 6\}$ & $\{16, 32, 64\}$ & $\{16, 32\}$ & -                & \\
                           & \multirow{2}{*}{RT}   & N & $\{2, 3, 4\}$ & $\{16, 32, 64\}$ & $\{16, 32\}$ & $\{16, 32, 64\}$ & \multirow{2}{*}{$\{1, 4, 16, 64, 256, 512\}$} \\
                           &                       & E & $\{2, 4, 6\}$ & $\{16, 32, 64\}$ & $\{16, 32\}$ & $\{16, 32, 64\}$ & \\
                           & \multirow{2}{*}{RT-S} & N & $\{2, 3, 4\}$ & $\{16, 32, 64\}$ & $\{16, 32\}$ & $\{16, 32, 64\}$ & \multirow{2}{*}{$\{1, 4, 16, 64, 256, 512\}$} \\
                           &                       & E & $\{2, 4, 6\}$ & $\{16, 32, 64\}$ & $\{16, 32\}$ & $\{16, 32, 64\}$ & \\
                           & \multirow{2}{*}{HCLT} & N & -             & $\{128, 256, 512, 1024\}$ & -            & -       & \multirow{2}{*}{$\{1, 4, 16, 64, 256, 512\}$} \\
                           &                       & E & -             & $\{128, 256, 512, 1024\}$ & -            & -       & \\
\bottomrule
\end{tabular}

        \end{small}
    \end{center}
\end{table*}

To implement the canonical orderings, we use the SciPy library \citep{virtanen2020scipy} for the BFT, DFT, and RCM variants and the RDKit library \citep{landrum2006rdkit} for the MCA variant.

\subsection{Baselines}
\label{sec:baselines}
We compare PGCs with various intractable graph DGMs. MoFlow \citep{zang2020moflow} is a one-shot, normalizing flow model, a composition of (several layers of) two types of invertible, affine transformations. The first one models nodes conditionally on observed edges and uses relational graph convolutional network (RGCN) \citep{schlichtkrull2018modeling} to parameterize the affine transformation. The second models edges and relies on a variant of the Glow model \citep{kingma2018glow} to parameterize the affine transformation. Continuous samples from the latent space are then mapped to discrete samples in the observation (graph) space, applying the dequantization (and quantization) to convert between discrete and continuous graphs (and vice versa). GraphAF \citep{shi2020graphaf} is an autoregressive, normalizing flow model that also uses an affine transformation parameterized by the RGCN to map continuous latent samples to discrete graphs, again using dequantization to perform the conversion from discrete to continuous samples. GraphDF \citep{luo2021graphdf} is an autoregressive, normalizing flow that eliminates the negative effect of dequantization and uses discrete modulo shift transformations to directly perform the mapping between the discrete latent space and discrete observation (graph) space.

EDP-GNN \citep{niu2020permutation}, GDSS \citep{jo2022score}, and DiGress \citep{vignac2023digress} are one-shot, diffusion models that noise and denoise input data through the forward and backward diffusion processes, receptively. The forward process of EDP-GNN perturbs input data with a sequence of increasing noise perturbations, jointly training a noise-conditioned neural network to estimate the score function---the gradient of the log distribution with respect to its input graph---by minimizing the score matching objective. The backward process utilizes annealed Langevin dynamics, recursively updating the score function with decreasing noise perturbations. GDSS realizes the forward (and backward) process through a system of positive (and negative) time-step stochastic differential equations, using the continuous-time version of the score-matching objective to train the node and edge score networks. EDP-GNN and GDSS rely on the dequantiazation, thus operating in the continuous space. DiGress is a discrete denoising diffusion model that works directly with the discrete node and edge attributes. The forward process relies on a Markov transition kernel to successively noise the node and edge attributes with discrete edits. In contrast, the backward process trains a graph transformer network to predict a clean graph from its noisy version, minimizing the cross entropy between the true and predicted graph. The simplicity and $\mathbb{S}_n$-equivariance of the denoising networks in these models is utilized to ensure that the targeted distribution over graphs is $\mathbb{S}_n$-invariant. GraphARM \citep{kong2023autoregressive} is a node-absorbing, autoregressive diffusion model that also operates directly in the discrete space. The forward process absorbs one node at each step, masking the node and the associated edges. This masking mechanism is repeated until all the nodes are absorbed and the graph becomes empty. The backward process recovers the input graph by jointly training a denoising network (which parametrizes the backward graph transition kernel) and a dedicated diffusion ordering network (which parameterizes a probability distribution over the graph ordering).

GraphEBM \citep{liu2021graphebm} is a one-shot, energy-based model which parametrizes the energy function with the RGCN, also relying on its $\mathbb{S}_n$-equivariance to make the resulting probability distribution over graphs $\mathbb{S}_n$-invariant. SPECTRE \citep{martinkus2022spectre} is a generative adversarial network that uses spectral decomposition to model a probability distribution over a graph conditionally on top-$k$ eigenvalues and eigenvectors.

\subsection{Computational Resources}
The experiments were conducted on a computational cluster with 56 Tesla A100 40GB GPUs. The jobs performing the gridsearch over the hyper-parameters in \cref{tab:hyper-parameters} were scheduled by SLURM 23.02.2. We limited each job to a single GPU. The computational time was restricted to 4 hours. All jobs were finished within this limit.

\section{Canonical Graph Orderings}
\label{sec:orderings}

\begin{figure*}[ht]
    \begin{center}
        \centerline{\input{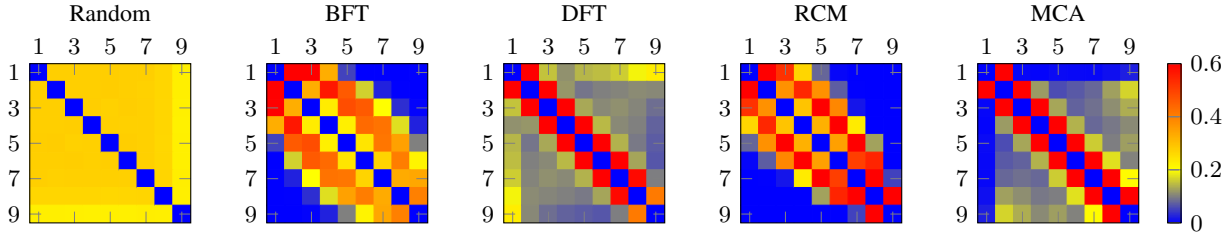}}
        \caption{\emph{The impact of different orderings on unnormalized empirical distribution over adjacency matrices.} The unnormalized empirical distribution is computed as the average of $\neg\mathbf{A}_{::1}$ for all graphs in the QM9 dataset \citep{ramakrishnan2014quantum}, where the entries of the matrix $\mathbf{A}_{::1}$ are equal to one if there is no edge between two nodes, and $\neg$ is the logical not. From left to right: all graphs in the dataset are randomly permuted (Random), ordered by the bread-first traversal (BFT), ordered by the deapth-first traversal (DFT), ordered by the reverse Cuthill-McKee (RCM) algorithm \citep{cuthill1969reducing}, ordered by the molecular canonicalization algorithm (MCA) which uses the domain knowledge to sort the graphs~\citep{schneider2015get}.
        }
        \label{fig:orderings}
    \end{center}
\end{figure*}

\cref{fig:orderings} illustrates a normalized empirical distribution over adjacency matrices of unordered and four canonically ordered graphs. For example, the Random case is nearly uniform, lacking any informative structure that could allow us to sample meaningful graphs. On the other hand, the BFT, DFT, RCM, and MCA orderings capture substantially more structural information, leading to a higher chance of generating more realistic and valid graph samples. The reason is that, for some orderings (e.g., BFT), the entries close to the diagonal are more concentrated, whereas those far from the diagonal are nearly zero. These concentrated patterns are also easier to learn than the Random case.

\section{Additional Results}
\label{sec:additional-results}

\textbf{Unconditional molecule generation.} Figures \ref{fig:unco-qm9-btree}-\ref{fig:unco-qm9-vtree} and Figures \ref{fig:unco-zinc250k-btree}-\ref{fig:unco-zinc250k-vtree} provide examples of generated molecular graphs for the QM9 and Zinc250k datasets, respectively. The graphs show each variant of the $\bm{\pi}$PGCs and each ordering from \cref{sec:orderings}. All the $\bm{\pi}$PGCs variants deliver very similar performance across all orderings for the QM9 dataset. However, for the Zinc250k dataset, we observe that the $\bm{\pi}$PGCs with the MCA and DFT orderings tend to generate rather elongated molecules that often contain no ring substructures. The BFT and RCM orderings, on the other hand, lead to more diverse and structured molecules with many ring substructures. The random ordering is more scattered and unusual compared to the other orderings.

\textbf{Conditional molecule generation.} Tables \ref{tab:generation-cond-qm9} and \ref{tab:generation-cond-zinc250k} display the molecular metrics for the conditional generation of molecules. In this experiment, we choose a specific molecular substructure to conditionally generate 10000 samples from the best models found in our original gridsearch (\cref{sec:pgc-variants}). To compute the molecular metrics, we take the original splits of the QM9 and Zinc250k datasets that were used to train the models, and for each of these splits, we select only the molecules that contain the particular substructure.

\begin{table*}[ht]
    \renewcommand{\arraystretch}{0.8}
    \caption{\emph{Conditional generation on the QM9 datasets}. The mean value of the molecular metrics for various implementations of the $\bm{\pi}$PGCs and different molecular subgraphs used to condition the model. The results are computed over five runs with different initial conditions. The \textcolor{c1}{1st}, \textcolor{c2}{2nd}, and \textcolor{c3}{3rd} best results are highlighted in colors. nAt and nBo are the average number of atoms and bonds, respectively, added to the newly generated part of the molecule. The parentheses in the Molecular scaffold column indicate the number of molecules in the training dataset containing this particular scaffold.}
    \label{tab:generation-cond-qm9}
    \begin{center}
        \begin{small}
            \begin{tabular}{llrrrrrrr}
\toprule
Molecular scaffold & Model & Valid$\uparrow$ & NSPDK$\downarrow$ & FCD$\downarrow$ & Unique$\uparrow$ & Novel$\uparrow$ & nAt & nBo \\\midrule
\multirow[c]{5}{*}{C1OCC=C1 (1295)} & BT & 81.05 & \color{c2} \textbf{0.125} & \color{c1} \textbf{11.11} & \color{c2} \textbf{9.99} & 99.95 & 3.80 & 6.31 \\
 & LT & 66.97 & \color{c1} \textbf{0.108} & \color{c2} \textbf{12.18} & \color{c1} \textbf{18.07} & \color{c3} \textbf{99.98} & 3.80 & 6.76 \\
 & RT & \color{c3} \textbf{89.56} & \color{c3} \textbf{0.181} & \color{c3} \textbf{12.84} & \color{c3} \textbf{3.17} & \color{c2} \textbf{100.00} & 3.80 & 6.72 \\
 & RT-S & \color{c2} \textbf{90.82} & 0.376 & 14.17 & 2.07 & 99.84 & 3.81 & 7.10 \\
 & HCLT & \color{c1} \textbf{91.34} & 0.424 & 13.27 & 0.28 & \color{c1} \textbf{100.00} & 1.43 & 1.47 \\
\cline{1-9}
\multirow[c]{5}{*}{N1NO1 (0)} & BT & \color{c2} \textbf{74.27} & - & - & \color{c2} \textbf{64.17} & 100.00 & 5.80 & 6.73 \\
 & LT & 45.28 & - & - & \color{c1} \textbf{83.45} & 100.00 & 5.80 & 7.63 \\
 & RT & \color{c3} \textbf{68.12} & - & - & \color{c3} \textbf{46.91} & \color{c3} \textbf{100.00} & 5.80 & 7.05 \\
 & RT-S & \color{c1} \textbf{82.13} & - & - & 39.42 & \color{c2} \textbf{100.00} & 5.80 & 6.96 \\
 & HCLT & 52.81 & - & - & 2.20 & \color{c1} \textbf{100.00} & 2.74 & 2.25 \\
\cline{1-9}
\multirow[c]{5}{*}{CCCO (59088)} & BT & \color{c3} \textbf{87.96} & \color{c3} \textbf{0.054} & \color{c3} \textbf{5.17} & \color{c2} \textbf{38.40} & \color{c3} \textbf{97.33} & 4.80 & 5.93 \\
 & LT & 73.10 & 0.064 & 6.24 & \color{c1} \textbf{48.26} & \color{c2} \textbf{97.99} & 4.79 & 5.86 \\
 & RT & \color{c2} \textbf{88.42} & \color{c1} \textbf{0.048} & \color{c2} \textbf{4.75} & \color{c3} \textbf{26.90} & 95.86 & 4.80 & 5.96 \\
 & RT-S & \color{c1} \textbf{94.42} & \color{c2} \textbf{0.052} & \color{c1} \textbf{4.68} & 19.29 & 94.48 & 4.80 & 5.88 \\
 & HCLT & 75.04 & 0.324 & 16.59 & 1.16 & \color{c1} \textbf{98.05} & 1.89 & 1.71 \\
\cline{1-9}
\multirow[c]{5}{*}{C1CNC1 (11421)} & BT & \color{c3} \textbf{85.09} & \color{c2} \textbf{0.032} & \color{c2} \textbf{3.95} & \color{c2} \textbf{37.59} & 99.99 & 4.80 & 7.37 \\
 & LT & 71.17 & \color{c1} \textbf{0.022} & 4.82 & \color{c1} \textbf{61.81} & \color{c2} \textbf{100.00} & 4.80 & 7.38 \\
 & RT & \color{c2} \textbf{92.63} & \color{c3} \textbf{0.034} & \color{c1} \textbf{3.65} & \color{c3} \textbf{23.39} & \color{c3} \textbf{99.99} & 4.80 & 7.15 \\
 & RT-S & \color{c1} \textbf{93.95} & 0.049 & \color{c3} \textbf{3.99} & 17.08 & 99.98 & 4.80 & 7.23 \\
 & HCLT & 75.50 & 0.299 & 15.08 & 1.18 & \color{c1} \textbf{100.00} & 1.87 & 1.69 \\
\cline{1-9}
\multirow[c]{5}{*}{CC(C)=O (11741)} & BT & \color{c3} \textbf{69.58} & \color{c2} \textbf{0.118} & \color{c2} \textbf{11.21} & \color{c2} \textbf{23.29} & 99.92 & 4.80 & 6.04 \\
 & LT & 54.26 & \color{c3} \textbf{0.127} & \color{c1} \textbf{10.88} & \color{c1} \textbf{31.96} & \color{c2} \textbf{99.93} & 4.80 & 6.28 \\
 & RT & \color{c2} \textbf{76.77} & 0.145 & 12.55 & \color{c3} \textbf{13.29} & \color{c3} \textbf{99.92} & 4.80 & 6.13 \\
 & RT-S & \color{c1} \textbf{80.83} & \color{c1} \textbf{0.112} & \color{c3} \textbf{11.32} & 10.17 & \color{c1} \textbf{99.94} & 4.80 & 6.27 \\
 & HCLT & 55.66 & 0.393 & 16.12 & 1.06 & 95.97 & 1.85 & 1.75 \\
\bottomrule
\end{tabular}

        \end{small}
    \end{center}
\end{table*}

\begin{figure*}[ht]
    \begin{center}
        \centerline{\includesvg[width=\linewidth]{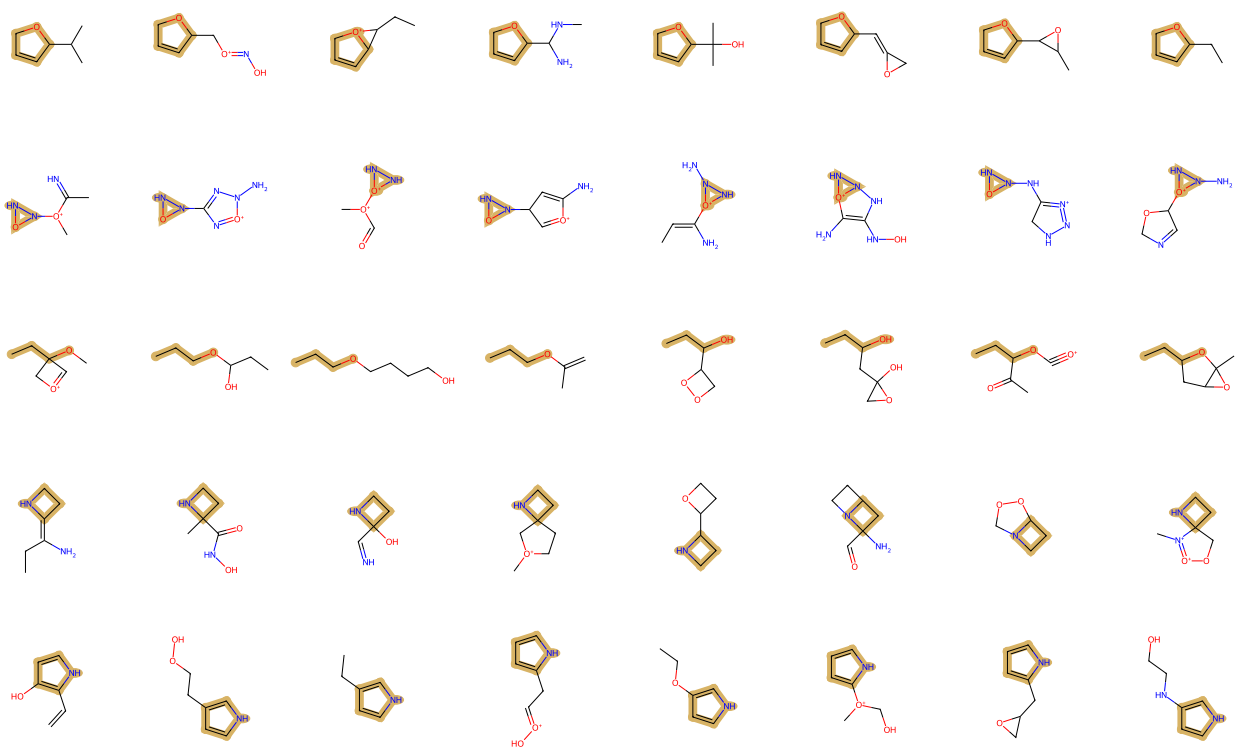}}
        \caption{\emph{Conditional generation on the QM9 dataset.} The \textcolor{c6}{yellow} area highlights the known part of the molecule. There is one such known part per row. Each column corresponds to a new molecule generated conditionally on the known part. The samples were produced by the RT-S variant of $\bm{\pi}$PGCs relying on the BFT ordering.}
        \label{fig:grid-cond-qm9}
    \end{center}
\end{figure*}

\begin{table*}[ht]
    \renewcommand{\arraystretch}{0.8}
    \caption{\emph{Conditional generation on the Zinc250k dataset}. The mean value of the molecular metrics for various implementations of the $\bm{\pi}$PGCs and different molecular subgraphs that are used to condition the model. The results are computed over five runs with different initial conditions. The \textcolor{c1}{1st}, \textcolor{c2}{2nd}, and \textcolor{c3}{3rd} best results are highlighted in colors. nAt and nBo are the average number of atoms and bonds, respectively, added to the newly generated part of the molecule. The parentheses in the Molecular scaffold column indicate the number of molecules in the training dataset containing this particular scaffold.}
    \label{tab:generation-cond-zinc250k}
    \begin{center}
        \begin{small}
            \begin{tabular}{llrrrrrrr}
\toprule
Molecular scaffold & Model & Valid$\uparrow$ & NSPDK$\downarrow$ & FCD$\downarrow$ & Unique$\uparrow$ & Novel$\uparrow$ & nAt & nBo \\\midrule
\multirow[c]{5}{*}{NS(=O)C1=CC=CC=C1 (0)} & BT & \color{c3} \textbf{24.45} & - & - & \color{c1} \textbf{98.83} & 100.00 & 14.17 & 16.22 \\
 & LT & 5.87 & - & - & \color{c2} \textbf{98.12} & 100.00 & 13.75 & 16.70 \\
 & RT & 16.18 & - & - & \color{c3} \textbf{93.53} & \color{c3} \textbf{100.00} & 13.85 & 16.24 \\
 & RT-S & \color{c2} \textbf{28.28} & - & - & 93.45 & \color{c2} \textbf{100.00} & 14.15 & 16.10 \\
 & HCLT & \color{c1} \textbf{76.23} & - & - & 3.43 & \color{c1} \textbf{100.00} & 1.22 & 1.23 \\
\cline{1-9}
\multirow[c]{5}{*}{CNC(C)=O (57329)} & BT & \color{c2} \textbf{17.19} & \color{c1} \textbf{0.101} & \color{c1} \textbf{32.52} & \color{c3} \textbf{99.91} & 100.00 & 18.16 & 20.60 \\
 & LT & 3.28 & \color{c3} \textbf{0.106} & 37.15 & \color{c2} \textbf{100.00} & 100.00 & 17.70 & 21.30 \\
 & RT & 6.53 & 0.106 & \color{c3} \textbf{34.54} & \color{c1} \textbf{100.00} & \color{c3} \textbf{100.00} & 17.78 & 20.57 \\
 & RT-S & \color{c3} \textbf{9.20} & \color{c2} \textbf{0.103} & \color{c2} \textbf{33.44} & 99.83 & \color{c2} \textbf{100.00} & 18.14 & 20.55 \\
 & HCLT & \color{c1} \textbf{24.50} & 0.365 & 44.17 & 21.97 & \color{c1} \textbf{100.00} & 3.45 & 2.69 \\
\cline{1-9}
\multirow[c]{5}{*}{O=C1CCCN1 (5287)} & BT & \color{c2} \textbf{20.66} & 0.093 & \color{c2} \textbf{31.76} & \color{c3} \textbf{99.68} & 100.00 & 17.16 & 19.59 \\
 & LT & 4.88 & \color{c1} \textbf{0.058} & 35.57 & \color{c1} \textbf{99.95} & 100.00 & 16.72 & 20.75 \\
 & RT & 7.94 & \color{c2} \textbf{0.062} & \color{c3} \textbf{32.59} & 99.43 & \color{c3} \textbf{100.00} & 16.84 & 19.86 \\
 & RT-S & \color{c3} \textbf{18.57} & \color{c3} \textbf{0.076} & \color{c1} \textbf{29.63} & \color{c2} \textbf{99.72} & \color{c2} \textbf{100.00} & 17.14 & 19.62 \\
 & HCLT & \color{c1} \textbf{75.59} & 0.431 & 45.75 & 9.90 & \color{c1} \textbf{100.00} & 3.50 & 3.50 \\
\cline{1-9}
\multirow[c]{5}{*}{C1CCNCC1 (28223)} & BT & \color{c2} \textbf{27.52} & 0.066 & \color{c2} \textbf{28.70} & \color{c2} \textbf{99.91} & 100.00 & 17.16 & 19.67 \\
 & LT & 7.45 & \color{c3} \textbf{0.064} & 32.42 & \color{c1} \textbf{100.00} & \color{c2} \textbf{100.00} & 16.71 & 19.66 \\
 & RT & 13.41 & \color{c1} \textbf{0.054} & \color{c3} \textbf{31.01} & 99.79 & \color{c3} \textbf{100.00} & 16.80 & 19.82 \\
 & RT-S & \color{c3} \textbf{25.47} & \color{c2} \textbf{0.060} & \color{c1} \textbf{27.89} & \color{c3} \textbf{99.88} & \color{c1} \textbf{100.00} & 17.14 & 20.28 \\
 & HCLT & \color{c1} \textbf{44.30} & 0.335 & 40.61 & 16.08 & 99.95 & 3.50 & 3.06 \\
\cline{1-9}
\multirow[c]{5}{*}{NS(=O)=O (15344)} & BT & \color{c2} \textbf{24.72} & \color{c3} \textbf{0.129} & \color{c2} \textbf{38.56} & \color{c3} \textbf{99.88} & 100.00 & 19.16 & 21.45 \\
 & LT & 3.16 & \color{c1} \textbf{0.120} & 47.55 & \color{c1} \textbf{100.00} & 100.00 & 18.69 & 22.45 \\
 & RT & \color{c3} \textbf{12.90} & \color{c2} \textbf{0.121} & \color{c3} \textbf{41.90} & \color{c2} \textbf{99.93} & \color{c3} \textbf{100.00} & 18.83 & 21.41 \\
 & RT-S & \color{c1} \textbf{31.29} & 0.145 & \color{c1} \textbf{36.74} & 98.60 & \color{c2} \textbf{100.00} & 19.15 & 21.26 \\
 & HCLT & 0.51 & 0.595 & 58.08 & 11.76 & \color{c1} \textbf{100.00} & 3.66 & 2.09 \\
\bottomrule
\end{tabular}

        \end{small}
    \end{center}
\end{table*}

\begin{figure*}[ht]
    \begin{center}
        \centerline{\includesvg[width=\linewidth]{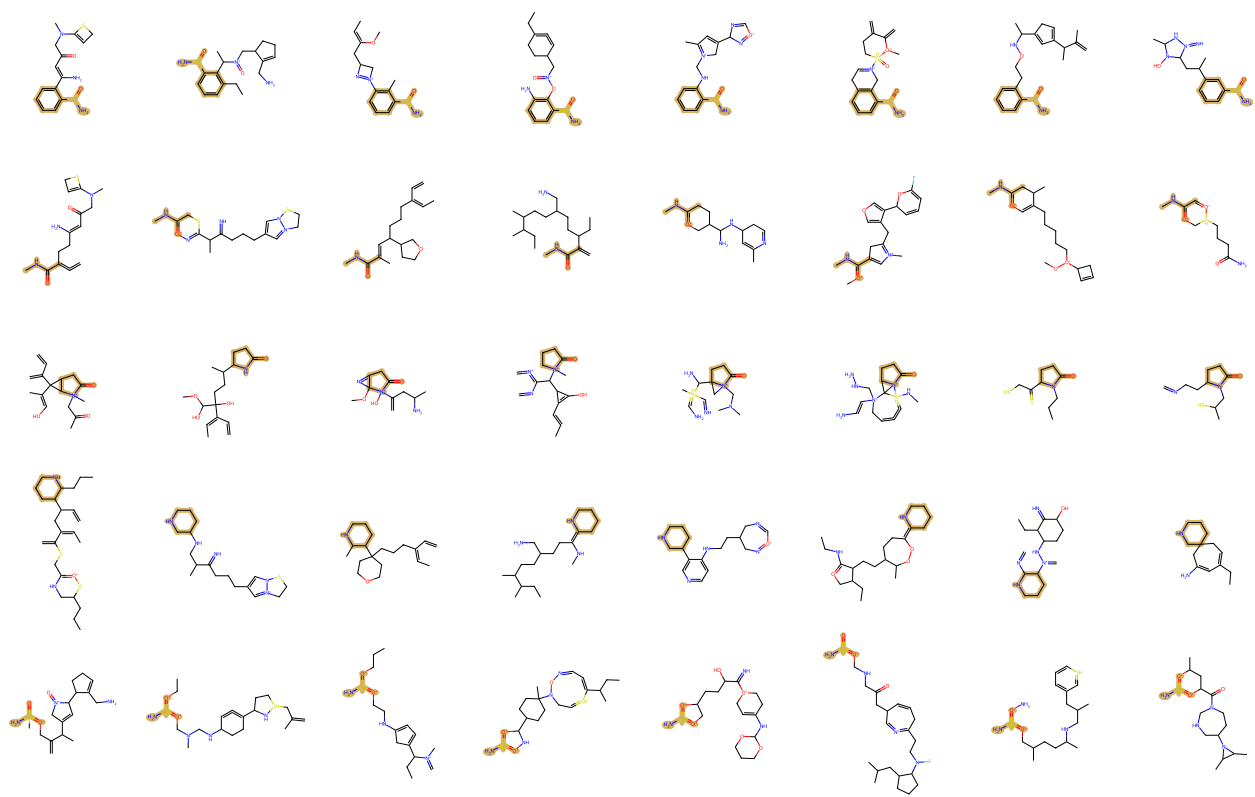}}
        \caption{\emph{Conditional generation on the Zinc250k dataset.} The \textcolor{c6}{yellow} area highlights the known part of the molecule. There is one such known part per row. Each column corresponds to a new molecule that is generated conditionally on the known part. The samples were produced by the RT-S variant of $\bm{\pi}$PGCs relying on the BFT ordering.}
        \label{fig:grid-cond-zinc250k}
    \end{center}
\end{figure*}

\begin{figure*}[ht]
    \begin{center}
        \centerline{\input{plots/orderings_performance_qm9.tikz}}
        \caption{\emph{$\bm{\pi}$PGCs for different orderings on the QM9 dataset.} The mean (bar) and standard deviation (error bounds) of the molecular metrics for various implementations of the $\bm{\pi}$PGCs that rely on the sorting to ensure the $\mathbb{S}_n$-invariance. The results are computed over five runs with different initial conditions.}
        \label{fig:ppgc-orderings-qm9}
    \end{center}
\end{figure*}

\begin{figure*}[ht]
    \begin{center}
        \centerline{\input{plots/orderings_performance_zinc250k.tikz}}
        \caption{\emph{$\bm{\pi}$PGCs for different orderings on the Zinc250k dataset.}  The mean (bar) and standard deviation (error bounds) of the molecular metrics for various implementations of the $\bm{\pi}$PGCs that rely on the sorting to ensure the $\mathbb{S}_n$-invariance. The results are computed over five runs with different initial conditions.}
        \label{fig:ppgc-orderings-zinc250k}
    \end{center}
\end{figure*}

\begin{figure*}[ht]
    \begin{center}
        \centerline{\input{plots/unconditional/qm9_marg_sort_btree_grid.tex}}
        \caption{\emph{Unconditional generation on the QM9 dataset.} Samples of molecular graphs for the BT variant of the $\bm{\pi}$PGCs and different canonical orderings presented in \cref{sec:orderings}. Invalid molecules were rejected during the sampling.}
        \label{fig:unco-qm9-btree}
    \end{center}
\end{figure*}

\begin{figure*}[ht]
    \begin{center}
        \centerline{\input{plots/unconditional/qm9_marg_sort_ctree_grid.tex}}
        \caption{\emph{Unconditional generation on the QM9 dataset.} Samples of molecular graphs for the HCLT variant of the $\bm{\pi}$PGCs and different canonical orderings presented in \cref{sec:orderings}. Invalid molecules were rejected during the sampling.}
        \label{fig:unco-qm9-ctree}
    \end{center}
\end{figure*}

\begin{figure*}[ht]
    \begin{center}
        \centerline{\input{plots/unconditional/qm9_marg_sort_ptree_grid.tex}}
        \caption{\emph{Unconditional generation on the QM9 dataset.} Samples of molecular graphs for the RT-S variant of the $\bm{\pi}$PGCs and different canonical orderings presented in \cref{sec:orderings}. Invalid molecules were rejected during the sampling.}
        \label{fig:unco-qm9-ptree}
    \end{center}
\end{figure*}

\begin{figure*}[ht]
    \begin{center}
        \centerline{\input{plots/unconditional/qm9_marg_sort_rtree_grid.tex}}
        \caption{\emph{Unconditional generation on the QM9 dataset.} Samples of molecular graphs for the RT variant of the $\bm{\pi}$PGCs and different canonical orderings presented in \cref{sec:orderings}. Invalid molecules were rejected during the sampling.}
        \label{fig:unco-qm9-rtree}
    \end{center}
\end{figure*}

\begin{figure*}[ht]
    \begin{center}
        \centerline{\input{plots/unconditional/qm9_marg_sort_vtree_grid.tex}}
        \caption{\emph{Unconditional generation on the QM9 dataset.} Samples of molecular graphs for the LT variant of the $\bm{\pi}$PGCs and different canonical orderings presented in \cref{sec:orderings}. Invalid molecules were rejected during the sampling.}
        \label{fig:unco-qm9-vtree}
    \end{center}
\end{figure*}

\begin{figure*}[ht]
    \begin{center}
        \centerline{\input{plots/unconditional/zinc250k_marg_sort_btree_grid.tex}}
        \caption{\emph{Unconditional generation on the Zinc250k dataset.} Samples of molecular graphs for the BT variant of the $\bm{\pi}$PGCs and different canonical orderings presented in \cref{sec:orderings}. Invalid molecules were rejected during the sampling.}
        \label{fig:unco-zinc250k-btree}
    \end{center}
\end{figure*}

\begin{figure*}[ht]
    \begin{center}
        \centerline{\input{plots/unconditional/zinc250k_marg_sort_ctree_grid.tex}}
        \caption{\emph{Unconditional generation on the Zinc250k dataset.} Samples of molecular graphs for the HCLT variant of the $\bm{\pi}$PGCs and different canonical orderings presented in \cref{sec:orderings}. Invalid molecules were rejected during the sampling.}
        \label{fig:unco-zinc250k-ctree}
    \end{center}
\end{figure*}

\begin{figure*}[ht]
    \begin{center}
        \centerline{\input{plots/unconditional/zinc250k_marg_sort_ptree_grid.tex}}
        \caption{\emph{Unconditional generation on the Zinc250k dataset.} Samples of molecular graphs for the RT-S variant of the $\bm{\pi}$PGCs and different canonical orderings presented in \cref{sec:orderings}. Invalid molecules were rejected during the sampling.}
        \label{fig:unco-zinc250k-ptree}
    \end{center}
\end{figure*}

\begin{figure*}[ht]
    \begin{center}
        \centerline{\input{plots/unconditional/zinc250k_marg_sort_rtree_grid.tex}}
        \caption{\emph{Unconditional generation on the Zinc250k dataset.} Samples of molecular graphs for the RT variant of the $\bm{\pi}$PGCs and different canonical orderings presented in \cref{sec:orderings}. Invalid molecules were rejected during the sampling.}
        \label{fig:unco-zinc250k-rtree}
    \end{center}
\end{figure*}

\begin{figure*}[ht]
    \begin{center}
        \centerline{\input{plots/unconditional/zinc250k_marg_sort_vtree_grid.tex}}
        \caption{\emph{Unconditional generation on the Zinc250k dataset.} Samples of molecular graphs for the LT variant of the $\bm{\pi}$PGCs and different canonical orderings presented in \cref{sec:orderings}. Invalid molecules were rejected during the sampling.}
        \label{fig:unco-zinc250k-vtree}
    \end{center}
\end{figure*}

\end{document}